\documentclass[11pt]{article}

\usepackage[preprint]{acl}

\usepackage{times}
\usepackage{latexsym}
\usepackage{booktabs}
\usepackage{subcaption}
\usepackage{multirow}
\usepackage{pifont}
\usepackage[table]{xcolor}

\newcommand{\xmark}{\ding{55}}
\usepackage[T1]{fontenc}

\usepackage[utf8]{inputenc}
\usepackage{amsmath}
\usepackage{amssymb}
\usepackage{stfloats}
\usepackage{cuted}
\usepackage{float}
\usepackage{placeins}
\usepackage{listings}
\usepackage{tcolorbox}
\usepackage{microtype}

\usepackage{inconsolata}

\usepackage{graphicx}
\definecolor{lightblue}{RGB}{173,216,230}
\definecolor{best}{RGB}{252,236,196}
\definecolor{second}{RGB}{223,235,253}
\usepackage{amsmath}
\usepackage{graphicx}

\title{
\raisebox{-1.5em}{\includegraphics[height=3em]{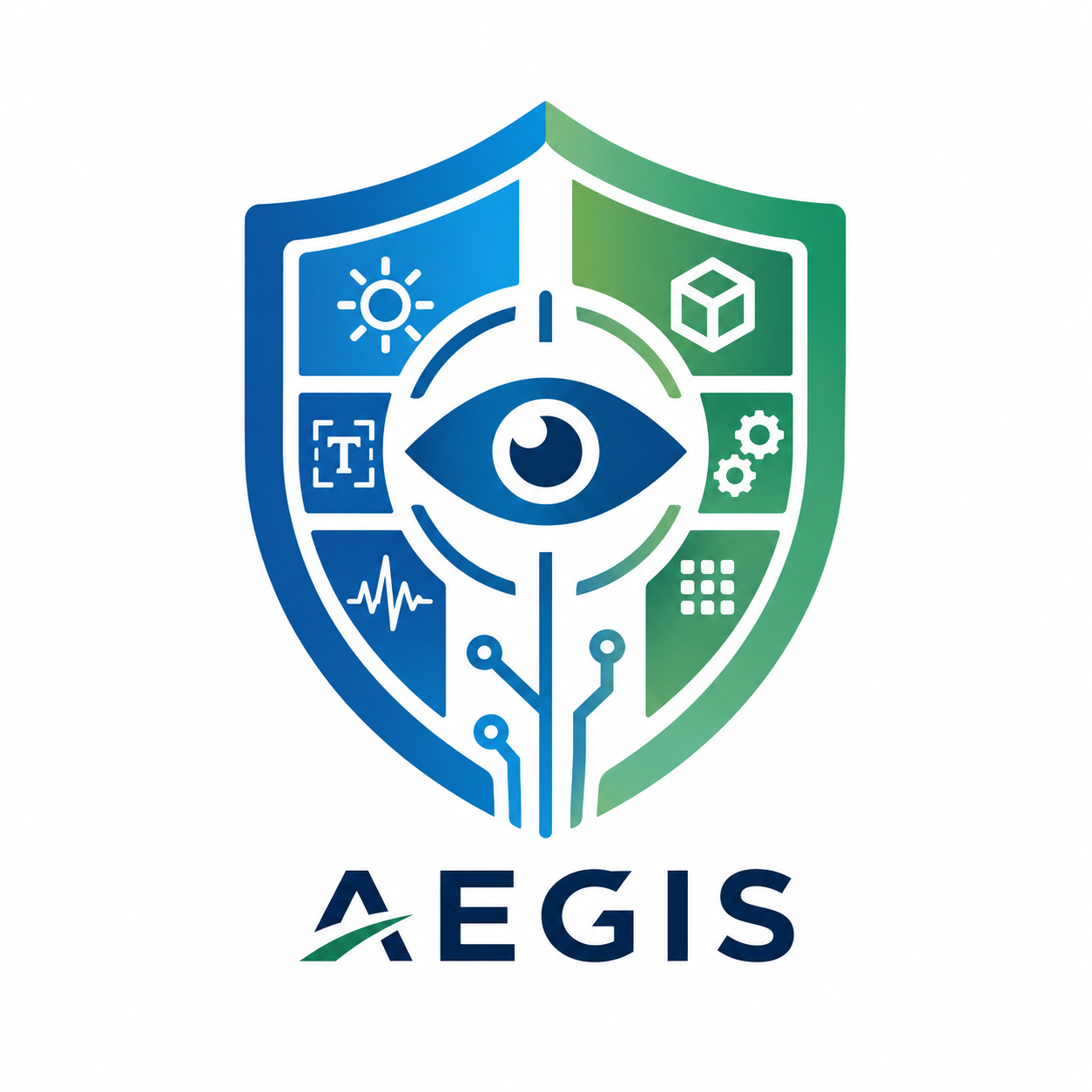}}
\hspace{-1em}
ClueAegis: Heuristic-to-Reasoning Cognitive-skill Learning for Unified Evidence-based Synthetic Image Detection
}

\author{
\textbf{Huangsen Cao\textsuperscript{1}},        \textbf{Hongkang Chu\textsuperscript{3}},
\textbf{Yuxi Li\textsuperscript{2}},
\textbf{Ying Zhang\textsuperscript{2}},\\
\textbf{Chen Li\textsuperscript{2}},
\textbf{Jing LYU\textsuperscript{2}},
\textbf{Yongwei Wang\textsuperscript{1}},
\textbf{Yu Zhao\textsuperscript{1}},
\textbf{Fei Wu\textsuperscript{1}}\\
\textsuperscript{1}Zhejiang University \\
\textsuperscript{2}WeChat Vision, Tencent Inc \\
\textsuperscript{3}University of the Chinese Academy of Sciences \\
\texttt{huangsen\_cao@zju.edu.cn} \\
}

\begin{document}
\maketitle
\begin{abstract}
The rapid advancement of generative models has made synthetic images increasingly realistic, challenging reliable detection. Existing methods are often limited to end-to-end classification or monolithic reasoning, and thus fail to model structured forensic reasoning and heterogeneous visual evidence. We revisit synthetic image detection from a cognitive perspective and propose a \textit{Heuristic-to-Reasoning} cognitive skill learning framework for evidence-based forensic analysis. Given an input image, our framework first extracts heuristic perceptual clues, selects the optimal forensic skill, and then performs skill-conditioned reasoning for evidence extraction and decision making. To support this paradigm, we introduce \textbf{ClueAegis-Bench}, which decomposes synthetic image detection into explicitly annotated forensic cognitive skills for structured evaluation beyond binary classification. Based on this benchmark, we propose \textbf{ClueAegis} (\underline{C}ognitive-skill \underline{L}earning for \underline{U}nified \underline{E}vidence-based Synthetic Image Detection), a two-stage agentic framework that conducts heuristic skill selection followed by evidence-guided reasoning through skill-conditioned toolchains. This design reformulates synthetic image detection as a configurable multi-skill reasoning process that bridges perception, skill selection, and forensic reasoning. Extensive experiments show that ClueAegis achieves state-of-the-art performance while improving cross-domain generalization and robustness. It also provides transparent reasoning trajectories and structured forensic evidence, offering a more explainable alternative to conventional end-to-end detectors.
\end{abstract}

\section{Introduction}

With the rapid advancement of generative artificial intelligence\cite{esser2024scaling,goodfellow2020generative,ho2020denoising}, high-quality AI-generated images have become increasingly prevalent across online platforms. While these images enrich visual content and enhance expressive diversity, they also blur the boundary between real and synthetic data, potentially facilitating misinformation dissemination and raising security concerns. Therefore, it is crucial to develop effective synthetic image detection methods that can fully leverage visual evidence for reliable discrimination.

\begin{figure}[t]
    \centering
    
    \includegraphics[width=\linewidth]{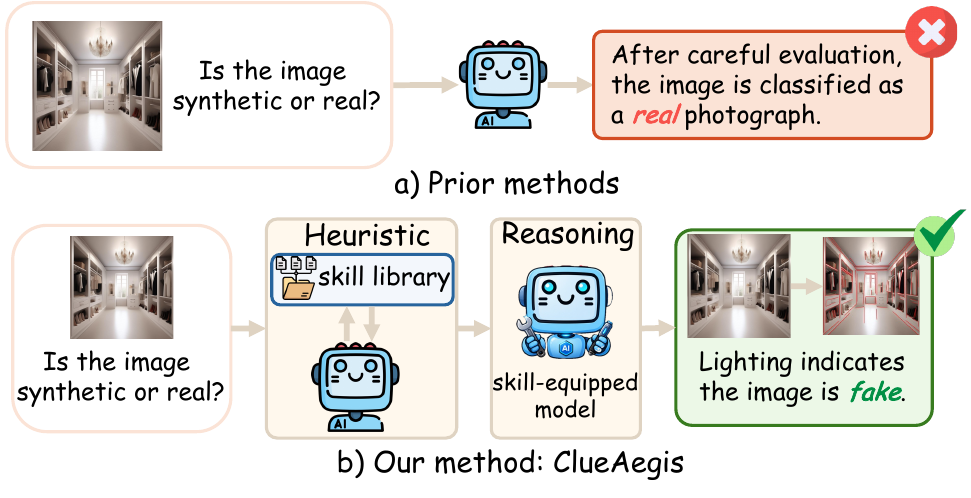}
    \vspace{-7mm}
\caption{(a) Prior synthetic image detection methods; (b) Our method: ClueAegis with a cognition-guided reasoning paradigm.}    
\label{fig:small_pipeline}
\vspace{-6mm}
\end{figure}

Recent advances in synthetic image detection~\cite{chai2020makes,liu2024forgery,ojha2023towards,wang2023dire} have achieved strong progress; however, most methods still rely on unified end-to-end or monolithic reasoning pipelines. While recent LLM-based approaches~\cite{li2025raidx,cao2025reveal,zhou2025aigi,xu2025fakeshield,huang2025sida,li2025fakebench,ye2025loki} introduce chain-of-thought reasoning for improved interpretability, they often treat heterogeneous forensic evidence as a single undifferentiated signal. However, forensic cues such as lighting, frequency artifacts, OCR errors, and anatomical inconsistencies exhibit fundamentally different distributions, and forcing them into a unified reasoning space can lead to feature entanglement and suboptimal evidence extraction. This suggests that the key limitation lies not only in reasoning depth, but also in the lack of structured decomposition of forensic cognition. Motivated by this, we propose a skill-driven decomposition with a heuristic-to-reasoning mechanism, which separates heterogeneous cues into structured skill spaces before performing higher-level reasoning (see Fig.~\ref{fig:small_pipeline}). This enables more faithful modeling of forensic workflows and improves both robustness and generalization.

One fundamental reason why existing methods struggle to learn such decomposed forensic cognition is the absence of heuristic-structured datasets and learning paradigms. Existing datasets~\cite{lu2023seeing,zhu2023genimage,jiang2025ivy} are typically organized according to generator types or data sources, rather than structured forensic heuristics and reasoning cues. Although several explainability benchmarks provide artifact-level descriptions, they fail to establish explicit correspondences between forensic evidence and reasoning processes, making it difficult to learn a decomposed heuristic-to-reasoning paradigm.

Therefore, the core limitations of existing methods can be summarized into two key issues: (i) \textbf{the lack of sufficiently large-scale and well-annotated datasets for fine-grained skill classification}, which hinders the effective learning of decomposed forensic cues and discriminative forensic skill representations; and (ii) \textbf{the absence of an explicit and principled heuristic-to-reasoning modeling paradigm}, which prevents models from adaptively transforming low-level perceptual forensic cues into conditional high-level reasoning processes during inference. Together, these limitations fundamentally restrict existing methods from achieving hierarchical cognitive modeling from fast heuristic perception to slow analytical reasoning.

To address these limitations, we introduce \textbf{ClueAegis-Bench}, a cognition-aware benchmark for synthetic image detection with explicit skill decomposition. Unlike conventional binary benchmarks, it adopts a skill-centric annotation scheme grounded in forensic reasoning, where each image is labeled with one of 12 predefined detection skills corresponding to its most discriminative forensic cue. This organizes heterogeneous artifacts into a unified skill space. Based on this decomposition, we reformulate detection as a structured reasoning process that explicitly separates perception, skill selection, and decision making. Models are required to perform perception-based analysis to select the appropriate forensic skill before making the final prediction. ClueAegis-Bench thus provides a testbed for studying skill-aware reasoning and cognitive decomposition in synthetic image detection, thereby improving generalization performance.

Based on ClueAegis-Bench, we propose \textbf{ClueAegis}, a two-stage heuristic-to-reasoning cognitive skill selection agent for synthetic image detection. In the first stage, the model operates as a perception-driven agent that interprets the input image and selects the most appropriate detection skill from a predefined skill space, thereby transforming cognitive understanding into an explicit actionable plan for downstream reasoning. In the second stage, the model performs skill-conditioned agentic detection by dynamically invoking corresponding analysis tools and feature transformation operators based on the selected skill. Each skill is associated with a distinct toolchain and processing paradigm, enabling an on-demand execution mechanism for structured forensic reasoning. This formulation reframes conventional static binary classification as a configurable, executable, and multi-skill agentic reasoning framework, thereby improving both adaptability and generalization capability in synthetic image detection.

Our contributions are summarized as follows:

\begin{itemize}
    \item \textbf{Task Re-formulation and Benchmark.} We reformulate synthetic image detection from binary classification to a skill-driven structured reasoning problem, and introduce ClueAegis-Bench to support this new paradigm with skill-oriented annotations.

    \item \textbf{Skill-driven Classification Space.} We organize forensic cues into 12 skill-oriented categories corresponding to different reasoning capabilities, forming a skill-driven classification space for adaptive forensic analysis.

    \item \textbf{Two-stage Forensic Reasoning.} We design a heuristic-to-reasoning two-stage framework that first builds global perceptual understanding and then performs cognition over relevant forensic skills for structured reasoning.

    \item \textbf{Empirical Results.} Experiments on multiple benchmarks show that our method achieves state-of-the-art performance, with improved generalization and robustness.
\end{itemize}

\section{Related Work}
\subsection{Synthetic Image Detection}

Early synthetic image detection methods mainly relied on CNN-based classifiers\cite{mo2018fake,nguyen2019capsule}, leveraging artifacts such as checkerboard patterns\cite{zhang2019detecting} and frequency-domain traces\cite{tan2024frequency}. With the emergence of pretrained vision models, CLIP-based approaches, such as UnivFD\cite{ojha2023towards}, achieved strong cross-domain generalization, while subsequent works further improved robustness through architectural and training refinements\cite{zhang2025towards,liu2024forgery,chen2025dual}. Other methods, including NPR\cite{tan2024rethinking}, HyperDet\cite{cao2024hyperdet},  AIDE\cite{yan2024sanity} and Patchcraft\cite{zhong2023patchcraft} enhanced detection by modeling pixel relationships and synthetic forensic traces. More recently, multimodal large language models have shifted the field toward reasoning-driven detection. Frameworks such as AIGI-Holmes\cite{zhou2025aigi}, REVEAL\cite{cao2025reveal}, Veritas\cite{tan2025veritas}, and Ivy-Fake\cite{jiang2025ivy} introduced interpretable and reasoning-based paradigms for generalized synthetic image forensic analysis.

\subsection{MLLM-Based Multi-Agent Frameworks}

Recent advances in large language models have enabled agentic frameworks that integrate task decomposition, tool invocation, and multi-step reasoning. Early multimodal agents, such as Visual ChatGPT\cite{wu2023visual}, MM-ReAct\cite{yang2023mm}, and HuggingGPT\cite{shen2023hugginggpt}, employed LLMs as planners for coordinating vision models and external tools, while frameworks like AutoGen\cite{wu2024autogen} and MetaGPT\cite{hong2024metagpt} further explored workflow-level agent collaboration. Recent MLLM studies extended this paradigm toward interleaved perception, reasoning, and action through iterative visual reasoning and tool-augmented interaction\cite{zheng2025deepeyes,hong2025deepeyesv2,yu2025introducing,zhou2025reinforced}. Meanwhile, skill-oriented agent learning methods, such as SkilRL\cite{xia2026skillrl} and Skill0\cite{lu2026skill0}, demonstrated that reusable cognitive skills can effectively bridge perception and decision making.

However, existing agentic MLLM frameworks mainly focus on general reasoning or tool use, while synthetic image detection lacks a structured forensic skill taxonomy and reasoning paradigm. ClueAegis addresses this by reformulating detection as a \textit{Heuristic-to-Reasoning} process, where heuristic-guided skill selection identifies forensic strategies, followed by evidence-based agentic reasoning for decision making.

\section{ClueAegis-Bench}
In cognitive psychology, dual-process theory distinguishes System 1—heuristic, intuitive perception—and System 2—deliberate analytical reasoning~\cite{kahneman2011thinking}. Building on this framework, we formulate synthetic-image detection as a structured forensic task, where System 1 performs rapid heuristic perception and System 2 conducts skill-grounded, evidence-driven reasoning.

\begin{figure}[t]
    \centering
    \includegraphics[width=\linewidth]{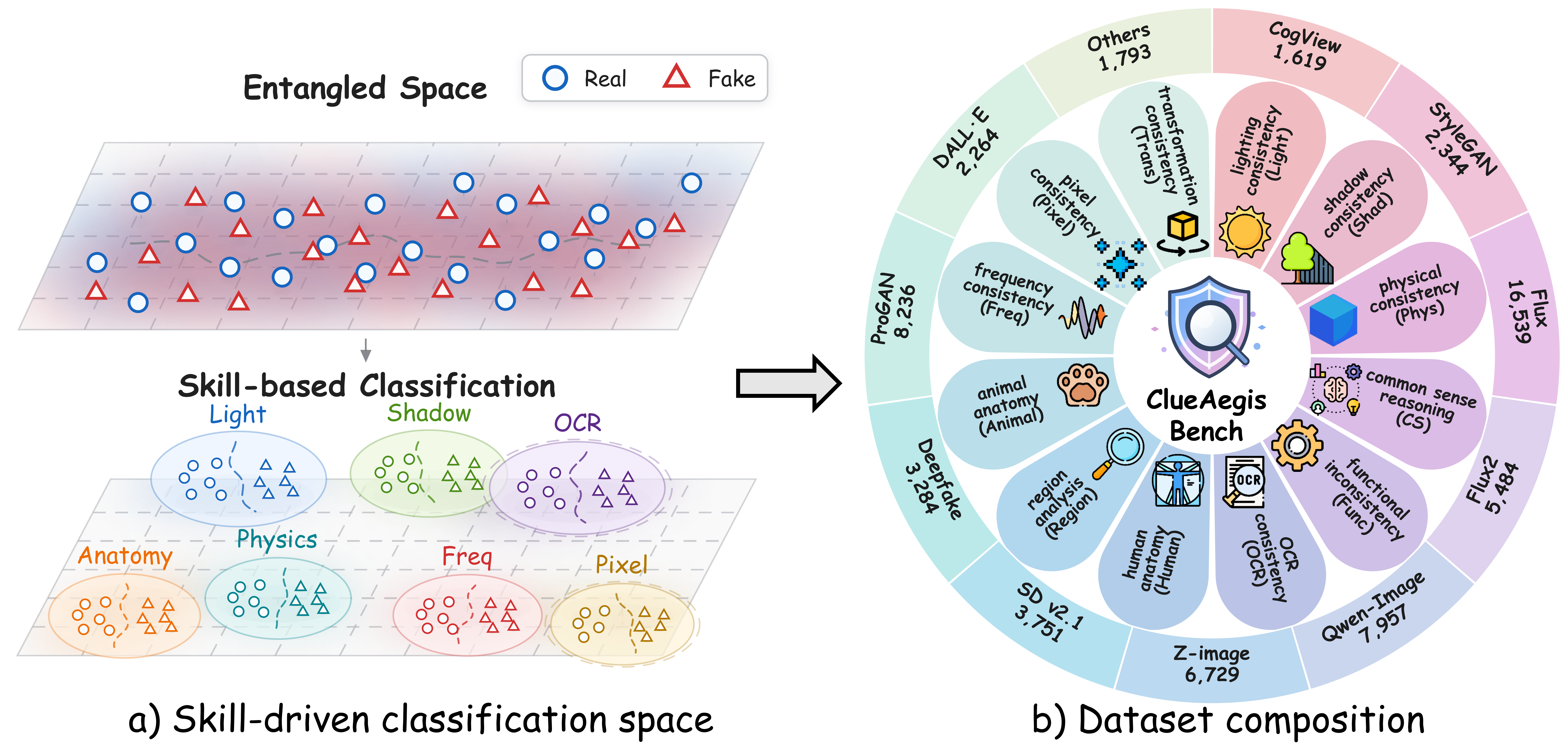}
\caption{Overview of ClueAegis-Bench. (a) Images are mapped from an entangled real/fake space into skill-specific classification  subspaces. (b) Skill-wise partitions across heterogeneous sources form ClueAegis-Bench with diverse synthetic distributions.}  
\label{fig:dataset_pie}
\vspace{-6mm}
\end{figure}
\subsection{ClueAegis-Bench Overview}
To support this paradigm, we construct a structured benchmark for skill-oriented synthetic image detection. 
As illustrated in Figure~\ref{fig:dataset_pie}, the task is formulated as a two-stage classification process: 
each image is first assigned to one of 12 predefined forensic skill categories, and then used to perform forensic authenticity discrimination conditioned on the predicted skill.

\textbf{Data Collection and Skill Annotation.}
We construct a large-scale dataset by aggregating approximately 6M images from multiple public sources, including Fake2M~\cite{lu2023seeing}, Projective-Geometry~\cite{sarkar2024shadows}, TGIF dataset~\cite{mareen2024tgif}, TextAtlas5M~\cite{wangtextatlas5m}, and WHOOPS~\cite{bitton2023breaking}. To further enhance the diversity of generative artifacts, we additionally synthesize 120K images using state-of-the-art generative models, including FLUX~\cite{black-forest-labs_flux_2024}, FLUX2~\cite{flux-2-2025}, Qwen-Image~\cite{wu2025qwen}, and Z-Image~\cite{cai2025z}. After data curation and balancing, we obtain a final benchmark consisting of 120K images in total, including balanced real and synthetic samples. Each forensic skill contains approximately 5K real and 5K synthetic images (See \textit{Appendix
\ref{visualization}}). 

For skill annotation, we define a set of 12 human-designed forensic skills that capture complementary types of authenticity cues, including \textit{lighting consistency (Light), shadow consistency (Shad), physical consistency (Phys), common-sense reasoning (CS), functional inconsistency (Func), OCR consistency (OCR), human anatomy (Human), region analysis (Region), animal anatomy (Animal), frequency consistency (Freq), pixel consistency (Pixel), and transformation consistency (Trans)}. Given a synthetic image, we employ \textbf{Qwen3-VL-235B-A22B}~\cite{bai2025qwen3} as an automatic multimodal forensic reasoner to evaluate whether each predefined skill can successfully detect forgery evidence in the image. Each skill is treated as an independent forensic detector. If Qwen3-VL determines that a specific skill reveals discriminative cues indicating that the image is synthetic, we retain the corresponding (image, skill) pair as a valid annotation. In this way, the dataset is constructed by linking fake images with the forensic skills that are most effective at exposing their artifacts, rather than performing conventional image-to-label assignment. For skills requiring external grounding signals, we integrate specialized tools; for example, OCR consistency is supported by PaddleOCR~\cite{cui2025paddleocr30technicalreport} for textual verification, among other task-specific tools (detailed in \textit{Appendix~\ref{details_skill}}). Further details on annotation protocols and skill definitions are provided in \textit{Appendix~\ref{details_bench} and ~\ref{details_skill}}.

\begin{figure*}[!t]
    \centering
    \includegraphics[width=\textwidth]{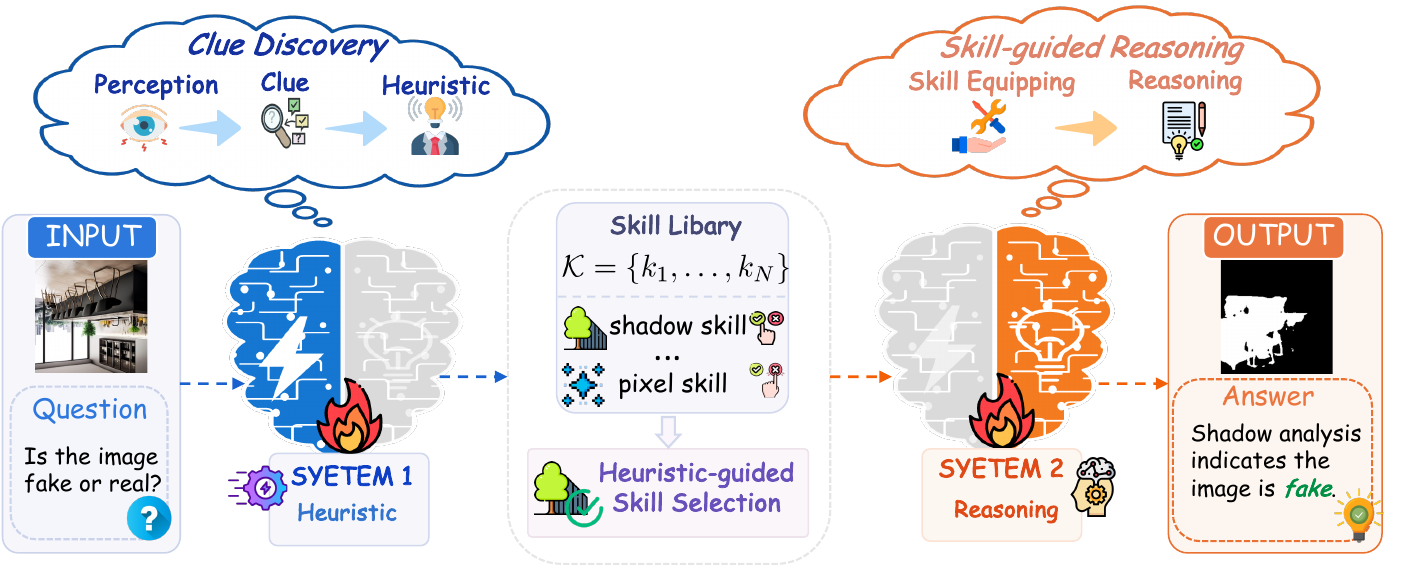}
    \caption{\textbf{Overview of ClueAegis.} A two-stage pipeline consisting of System 1 (heuristic) and System 2 (reasoning).}
    \label{fig:pipline}
    \vspace{-4mm}
\end{figure*}

\section{Method}
In this section, we present the heuristic--to-reasoning framework of ClueAegis.
\subsection{Overview of ClueAegis}
Inspired by the dual-process theory of human cognition, we propose ClueAegis, a synthetic image detection framework based on a heuristic-to-reasoning cognitive paradigm. As illustrated in Figure~\ref{fig:pipline}, ClueAegis reformulates conventional end-to-end detection as a cognitive forensic reasoning process with two stages: \textbf{Heuristic-guided Skill Selection} and \textbf{Skill-guided Forensic Reasoning}. 

\textbf{Heuristic-guided Skill Selection.}
Traditional synthetic image detection methods typically rely on unified feature representations and a single-stage binary classification framework. However, forgery artifacts exhibit substantial heterogeneity across semantic domains, physical structures, and generation processes, resulting in highly entangled forensic distributions. Such diversity makes globally shared decision boundaries difficult to generalize across real-world synthetic images.

To address this issue, we reformulate synthetic image detection as a skill-conditioned forensic reasoning problem by decomposing authenticity analysis into multiple forensic cognitive skills. Based on this formulation, we construct a skill-conditioned space for adaptive reasoning.

Given an input image $x$, perceptual forensic evidence is first extracted through a visual encoder.

\begin{equation}
c = f_{\theta}(x),
\end{equation}
where $f_{\theta}$ denotes the cognitive perception encoder and $c$ represents forensic clues composed of abnormal visual patterns, structural inconsistencies, and latent semantic conflicts.

We define a forensic cognitive skill library:
\begin{equation}
\mathcal{K}=\{k_1,k_2,\dots,k_N\},
\end{equation}
where each skill corresponds to a specialized forensic reasoning capability, such as OCR verification, frequency analysis, physical consistency reasoning, or anatomical analysis, among others (see \textit{Appendix~\ref{details_skill}} for full definitions).

To determine the most relevant forensic reasoning pathway, the model selects a skill from the predefined skill library conditioned on the extracted forensic evidence:

\begin{equation}
s^{*}
=
\mathrm{VLLM\text{-}Select}(c, \mathcal{K}),
\end{equation}
where selection is performed by prompting a large language model to match forensic evidence with the most relevant cognitive skill in the skill library.

The selected skill then guides subsequent forensic reasoning for authenticity prediction, forming a skill-conditioned inference process, which can be viewed as a System~1-like heuristic routing stage that rapidly identifies the most relevant forensic skill.

\textbf{Skill-guided Forensic Reasoning.}
After selecting the forensic skill $s^{*}$, ClueAegis performs skill-guided forensic reasoning for evidence-driven authenticity analysis. Specifically, the selected skill is transformed into a specialized forensic prompt that guides the large language model to focus on corresponding anomaly patterns and forensic cues.

Conditioned on the selected skill, the reasoning process is formulated as:

\begin{equation}
\mathcal{R}
=
P_{\mathcal{L}}(x,s^{*}),
\end{equation}
where $P_{\mathcal{L}}$ denotes the skill-conditioned forensic reasoning process and $\mathcal{R}$ represents the generated forensic reasoning trajectory.

During reasoning, the model may invoke external forensic tools and APIs for specialized evidence verification:

\begin{equation}
o = \mathcal{A}(x,s^{*},\mathcal{R}),
\end{equation}
where $\mathcal{A}$ denotes the forensic agent system and $o$ represents auxiliary forensic observations returned by external tools.

The final authenticity prediction is obtained by integrating the reasoning trajectory and external forensic evidence:
\begin{equation}
y=\Phi(\mathcal{R},o),
\end{equation}
where $\Phi(\cdot)$ denotes the forensic decision function that maps the reasoning trajectory and external evidence to a binary authenticity prediction.

From a cognitive perspective, this process resembles the analytical reasoning mechanism of System~2, where humans perform deliberate evidence verification and progressively refine forensic judgments under task-specific cognitive skills.

Overall, previous methods typically model synthetic image detection as
$p(y \mid x, \mathcal{E})$, where $\mathcal{E}=\{e_i\}_{i=1}^{N}$ denotes a set of entangled forensic evidence and prediction is made through joint reasoning over all cues. In contrast, ClueAegis reformulates the problem as
$p(y \mid x, e^{*})$, where $e^{*}$ is the most relevant skill-induced forensic clue selected from a predefined set of cognitive skills, enabling focused reasoning on a single discriminative evidence source.

\subsection{Two-Stage Cognitive Training}

Our two-stage design models the cognitive inference process rather than the optimization pipeline. Specifically, ClueAegis decomposes synthetic image detection into two sequential stages: (1) heuristic-guided forensic skill selection, and (2) skill-guided reasoning. To learn this behavior, we employ supervised fine-tuning (SFT) followed by group relative policy optimization (GRPO).

\textbf{Stage I: Heuristic Skill Learning.}
In the first stage, the model learns to analyze forensic evidence and select the most relevant forensic skill for a given image. Given an input image $x$, the model predicts the corresponding forensic skill $s^{*}$:

\begin{equation}
\mathcal{L}_{\text{skill}}
=
-\log P_{\theta}(s^{*}\mid x).
\end{equation}

This stage establishes the mapping between heuristic forensic evidence and the predefined forensic skill space.

\textbf{Stage II: Skill-guided Forensic Reasoning.}
In the second stage, the model performs authenticity analysis conditioned on the selected forensic skill. Given $x$ and $s^{*}$, the model conducts skill-specific reasoning and predicts the final authenticity label $y$:

\begin{equation}
\mathcal{L}_{\text{det}}
=
-\log P_{\theta}(y\mid x,s^{*}).
\end{equation}

This stage enables the model to adapt its forensic reasoning behavior according to different cognitive skills.

\textbf{GRPO-based Cognitive Alignment.}
To improve reasoning consistency and robustness, we apply GRPO for preference optimization over forensic outputs. The reward function jointly considers answer correctness, skill selection consistency, and format compliance:

\begin{equation}
R
=
\lambda_1 R_{\text{ans}}
+
\lambda_2 R_{\text{skill}}
+
\lambda_3 R_{\text{format}}.
\end{equation}

Here, $R_{\text{ans}}$ measures prediction correctness against ground truth, $R_{\text{skill}}$ checks whether the selected forensic skill matches the annotation, and $R_{\text{format}}$ enforces the required two-stage reasoning structure (skill selection followed by prediction).

Given sampled outputs $\{o_1,\dots,o_G\}$, the GRPO objective is:

\begin{equation}
\mathcal{L}_{\text{GRPO}}
=
-
\mathbb{E}_{o\sim\pi_{\theta}}
\left[
\frac{R(o)-\mu_R}{\sigma_R}
\log\pi_{\theta}(o\mid x)
\right],
\end{equation}
where $\mu_R$ and $\sigma_R$ are group-wise statistics over $\{o_1,\dots,o_G\}$. This encourages stable preference learning and structured reasoning behavior.

\section{Experiment}
\begin{table*}[htbp]
    \centering
    \caption{\textbf{Performance on ClueAegis-Bench.} ClueAegis achieves the best overall performance and surpasses the previous state-of-the-art method by \textbf{6.99\%} in accuracy, demonstrating superior generalization capability across diverse synthetic image distributions. \colorbox{best}{\textbf{Best}} and \colorbox{second}{\underline{second-best}} are highlighted.
    }

    \label{tab:AEGIS-Bench}
    \footnotesize
    \renewcommand\arraystretch{1.15}

    \resizebox{\textwidth}{!}{
    \begin{tabular}{c ccccccccccccc c}
        \toprule
        \multirow{2}{*}{\textbf{Method}} 
        & \multicolumn{12}{c}{\textbf{Forensic Consistency Metrics}} 
        & \multirow{2}{*}{\textbf{Avg.}} \\
        
        \cmidrule(lr){2-13}
        
        & \textbf{Light} 
        & \textbf{Shadow} 
        & \textbf{Phys} 
        & \textbf{CS} 
        & \textbf{Func} 
        & \textbf{OCR} 
        & \textbf{Human} 
        & \textbf{Region} 
        & \textbf{Animal} 
        & \textbf{Freq} 
        & \textbf{Pixel} 
        & \textbf{Trans} \\
        
        \midrule
        CNNSpot & \cellcolor{second}\underline{94.15} & \cellcolor{second}\underline{93.00} & 93.55 & 94.35 & \cellcolor{second}\underline{94.30} & 93.15 & 93.50 & 63.80 & 93.85 & 64.00 & 93.00 & 85.80 & 88.04 \\
        FreqNet & 80.10 & 79.50 & 89.95 & 93.70 & 92.50 & 87.15 & 92.70 & 58.50 & 90.60 & 87.10 & 88.40 & 74.20 & 84.53 \\
        UnivFD & 93.50 & 87.25 & 90.90 & 92.70 & 92.00 & 90.35 & 92.70 & 65.15 & 92.10 & 87.30 & 82.85 & 75.20 & 86.83 \\
        NPR & 82.60 & 75.70 & 87.20 & 88.75 & 88.20 & 82.70 & 87.40 & 81.95 & 87.50 & 88.40 & 85.20 & 82.70 & 84.86 \\
        HyperDet & 50.30 & 50.10 & 86.90 & 89.85 & 86.60 & 50.30 & 90.60 & 50.20 & 90.35 & 86.75 & 80.25 & 68.80 & 73.42 \\
        AIDE & 93.80 & 90.65 & \cellcolor{second}\underline{94.30} & \cellcolor{second}\underline{94.55} & \cellcolor{second}\underline{94.30} & 93.00 & 94.55 & 64.85 & 94.25 & 93.50 & 92.70 & 83.45 & 90.33 \\
        DDA & 94.00 & 92.70 & 94.55 & 94.65 & 94.25 & \cellcolor{second}\underline{93.35} & 94.60 & 73.10 & 94.60 & 94.35 & 93.60 & \cellcolor{second}\underline{90.45} & \cellcolor{second}\underline{92.02} \\
        AIGI-Holmes & 51.75 & 51.05 & 87.75 & 91.15 & 83.40 & 91.15 & 91.45 & \cellcolor{second}\underline{84.85} & 83.50 & \cellcolor{second}\underline{98.95} & \cellcolor{second}\underline{97.75} & 83.40 & 83.01 \\
        REVEAL & 87.25 & 85.35 & 80.55 & 91.25 & 91.30 & 92.60 & 91.55 & 80.45 & 93.65 & 74.45 & 72.95 & 61.35 & 83.56 \\
        Veritas & 88.35 & 82.45 & 76.35 & 81.70 & 76.05 & 63.85 & 86.75 & 71.25 & 80.25 & 79.60 & 75.15 & 65.60 & 77.28 \\
        Ivy-xDetector & 50.65 & 50.95 & 92.95 & 94.75 & 91.25 & 72.95 & \cellcolor{second}\underline{94.70} & 80.55 & \cellcolor{second}\underline{94.95} & 78.65 & 89.75 & 64.30 & 79.70 \\
        \textit{\textbf{ClueAegis}} & \cellcolor{best}\textbf{99.80} & \cellcolor{best}\textbf{99.70} & \cellcolor{best}\textbf{99.40} & \cellcolor{best}\textbf{99.75} & \cellcolor{best}\textbf{99.75} & \cellcolor{best}\textbf{99.85} & \cellcolor{best}\textbf{99.70} & \cellcolor{best}\textbf{95.50} & \cellcolor{best}\textbf{99.75} & \cellcolor{best}\textbf{99.95} & \cellcolor{best}\textbf{99.55} & \cellcolor{best}\textbf{95.40} & \cellcolor{best}\textbf{99.01} \\
        \bottomrule
    \end{tabular}
    }
\end{table*}

\subsection{Experimental Settings}
To comprehensively evaluate ClueAegis, we adopt Qwen3.5-9B as the backbone model and conduct all experiments based on this unified configuration. We train the model on ClueAegis-Bench (96K samples) and evaluate on three primary datasets: ClueAegis-Bench, the first skill-oriented dataset; AIGI-Now~\cite{chen2025task}, a dataset built with recent state-of-the-art generators; and GenImage~\cite{zhu2023genimage}, a large-scale dataset with over one million samples. We further include evaluations on six additional datasets in Section \ref{additional} (training details in \textit{Appendix \ref{training_details}}).

\paragraph{Baselines.}We compare ClueAegis with a range of state-of-the-art synthetic image detection methods, including CNNSpot\cite{wang2020cnn}, FreqNet\cite{tan2024frequency}, UnivFD\cite{ojha2023towards}, NPR\cite{tan2024rethinking}, HyperDet\cite{cao2024hyperdet}, AIDE\cite{yan2024sanity}, DDA\cite{chen2025dual}, AIGI-Holmes\cite{zhou2025aigi}, REVEAL\cite{cao2025reveal}, Veritas\cite{tan2025veritas} and Ivy-xDetector\cite{jiang2025ivy} as representative baselines.

\paragraph{Evaluation Metrics.}Following prior work in synthetic image detection, we report Accuracy (ACC) as the primary metric in the main paper, with F1-score results provided in Appendix~\ref{F1}. For ClueAegis and LLM-based baselines, we map textual outputs (i.e., \textit{Real} or \textit{Fake}) into binary predictions for evaluation. For non-LLM methods, we follow their official decision thresholds. We do not report AP since several reasoning-based methods produce textual outputs without calibrated confidence scores, making AP not directly applicable under a unified setting. Additional results, including few-shot performance evaluation and qualitative results, are provided in \textit{Appendix~\ref{app:few_shot} and \ref{Qualitative}}.
\subsection{Comparison to State-of-the-Art Detectors}

\paragraph{Evaluation on ClueAegis-Bench.} Table~\ref{tab:AEGIS-Bench} presents a comparison between ClueAegis and existing synthetic image detection methods on \textit{ClueAegis-Bench}. ClueAegis achieves the best overall performance, surpassing the previous state-of-the-art method by \textbf{6.99\%} in accuracy. The results demonstrate that ClueAegis effectively learns skill-driven forensic representations, enabling more accurate authenticity reasoning.

\begin{table*}[htbp]
\centering
\caption{\textbf{Generalization on AIGI-Now.} ClueAegis achieves the best overall performance in terms of accuracy, demonstrating superior generalization capability across advanced generative models.}
\label{tab:aigi-now}
\small
\resizebox{\textwidth}{!}{
\begin{tabular}{l|cc|cc|cc|cc|cc|cc|cc|cc|cc|c}
\toprule
\multirow{2}{*}{\textbf{Detector}} & 
\multicolumn{2}{c|}{\textbf{FLUX-dev}} & 
\multicolumn{2}{c|}{\textbf{FLUX-kera}} & 
\multicolumn{2}{c|}{\textbf{FLUX-kontext}} & 
\multicolumn{2}{c|}{\textbf{FLUX-pro}} & 
\multicolumn{2}{c|}{\textbf{gpt4o}} & 
\multicolumn{2}{c|}{\textbf{jimeng}} & 
\multicolumn{2}{c|}{\textbf{keling}} & 
\multicolumn{2}{c|}{\textbf{minimax}} & 
\multicolumn{2}{c|}{\textbf{Nano}} & 
\multirow{2}{*}{\textbf{Avg.}} \\
& pix & sem & pix & sem & pix & sem & pix & sem & pix & sem & pix & sem & pix & sem & pix & sem & pix & sem & \\

\midrule
\multicolumn{20}{c}{\textbf{Conventional Binary Detection Methods}} \\
CNNSpot & 76.90 &74.85  &78.30  &59.05  &62.70  &57.35  &64.80  &63.95  &74.90  &61.00  &68.75  &60.65  &75.00  &72.35  &72.40  &57.65  &76.45  &70.75  &  68.21\\
FreqNet & 72.80 &52.15  &72.50  &51.35  &60.15  &53.45  &49.05  &50.30  &74.80  &55.50  &56.30  &52.35  &73.85  &53.65  &68.55  &52.25  &76.20  &50.05  & 59.74 \\

UnivFD & 85.20 & 89.75 & 77.70 &85.40  &62.05  &73.10  &61.90  &\cellcolor{second}\underline{87.55}  &86.25  &91.85  &70.25  &\cellcolor{second}\underline{91.75}  & 85.05 & \cellcolor{second}\underline{90.60} &68.10  &80.75  &83.95  &\cellcolor{second}\underline{92.40}  & 81.31 \\

NPR & 54.85 & 48.05 &55.85  &46.15  &52.80  &44.75  &48.45  &48.15  &54.20  &49.40  &51.10  &43.95  &54.20  &63.00  &53.75  &44.50  &55.20  &55.15  & 51.31 \\

AIDE & 87.70 &72.50  &86.85  &61.90  &76.20  & 61.50 &71.25  &70.55  &86.25  &73.40  &71.15  &71.30  &87.10  &79.30  &81.60  &63.55  &88.95  &86.55  &76.53  \\
DDA & 78.30 &89.70  &78.45  &65.90  &73.90  &64.50  &63.10  &86.20  &78.00  &86.15  &60.85  &81.45  &78.35  &87.30  &75.95  &65.85  &79.45  &90.05  & 76.86 \\
\midrule
\multicolumn{20}{c}{\textbf{MLLM-based Detection Methods}} \\
AIGI-Holmes & \cellcolor{second}\underline{98.70} &85.30  & 90.60 &74.80  &\cellcolor{second}\underline{88.40}  &74.15  &\cellcolor{best}\textbf{99.00}  &80.45  &\cellcolor{best}\textbf{99.65}  &79.60  &\cellcolor{best}\textbf{99.60}  &85.00  &97.50  &87.20  &88.05  &65.90  &\cellcolor{best}\textbf{96.55}  &\cellcolor{best}\textbf{96.15}  & 88.14 \\
REVEAL & 92.50 & \cellcolor{second}\underline{91.00} & 94.00 &84.65  & 87.65 & \cellcolor{second}\underline{88.00} & 79.95 & 74.55 &  92.35 &  87.40 &  91.15 & 90.00 & 93.65 &88.80  &66.95  &75.65  &80.45  &83.35  & 85.67 \\
Veritas & 84.70 &81.35  &68.05  &75.40  &69.20  &73.85  &60.75  &77.40  &69.15  &80.70  &68.10  &81.40  &84.70  &82.80  &60.25  &73.80  &81.65  &83.20  & 75.36 \\
Ivy-xDetector & 95.40 &88.35  &\cellcolor{second}\underline{94.80}  &\cellcolor{second}\underline{86.85}  & 85.45 & 81.30 &90.25  &87.45  &95.45  &87.75  &\cellcolor{second}\underline{95.65}  &87.90  &\cellcolor{second}\underline{95.55}  &87.80  &\cellcolor{second}\underline{92.50}  &\cellcolor{second}\underline{86.80}  &\cellcolor{second}\underline{95.30 } &87.50  & \cellcolor{second}\underline{90.11} \\
\textit{\textbf{ClueAegis}} & \cellcolor{best}\textbf{98.85}  &\cellcolor{best}\textbf{99.75}  &\cellcolor{best}\textbf{98.70}  &\cellcolor{best}\textbf{99.45}  &\cellcolor{best}\textbf{96.20}  & \cellcolor{best}\textbf{98.90} & \cellcolor{second}\underline{90.60} &\cellcolor{best}\textbf{99.20}  &98.75  &\cellcolor{best}\textbf{99.50}  & 92.05 &\cellcolor{best}\textbf{99.15}  &\cellcolor{best}\textbf{98.35}  &\cellcolor{best}\textbf{99.15}  &\cellcolor{best}\textbf{97.90}  &\cellcolor{best}\textbf{99.45}  &83.95  &\cellcolor{second}\underline{92.40}  &  \cellcolor{best}\textbf{97.99}\\

\bottomrule
\end{tabular}
}
\end{table*}
\begin{table*}[htbp]
\caption{\textbf{Performance on GenImage.} ClueAegis achieves the best overall performance on the GenImage benchmark, surpassing existing methods by \textbf{0.67\%} in accuracy and demonstrating strong generalization capability on conventional synthetic image detection settings.}
\vspace{-5mm}

\begin{center}
\resizebox{\textwidth}{!}{
\begin{tabular}{lccccccccc}
\toprule
Method    &{Midjourney} &{SD v1.4} & {SD v1.5} & {ADM} &{GLIDE} &{Wukong} &{VQDM} &{BigGAN} & {{Avg.}} \\ 
\midrule

{CNNSpot}     & 75.46 & 70.31 & 69.24  & 60.10 & 68.39 & 69.97 & 49.83  & 45.45 & 63.59\\ 
{FreqNet}      & 66.70 & 62.45 & 60.90  & 41.08 & 58.81 & 61.14 & 53.07  & 47.97 & 56.52 \\ 
{UnivFD}    & 76.25 & 75.36 & 73.34  & 58.91 & 63.46 & 73.20 & 63.74  & 57.52 & 67.72 \\ 
{NPR}   & 60.79 & 61.26 & 58.40  & 61.73 & 59.42 & 60.29 & 57.96  & 50.33 & 58.77 \\
{HyperDet}   & 45.61 & 56.53 & 53.10  & 62.12 & 37.44 & 61.60 & 60.43  & 33.94 & 51.35 \\

{AIDE}   & 82.23 & 79.34 & 79.44  & 73.94 & 82.99 & 78.66 & 74.49  & 66.09 & 77.15 \\
{DDA}   & 80.64 & 79.73 & 78.74  & 75.00 & 76.34 & 78.43 & 79.23  & 79.62 & 78.47\\
{AIGI-Holmes}     & 86.10 & \cellcolor{second}\underline{93.17} & 91.22  & 84.32 & 72.53 & 92.10 & 89.77  & 91.00 & 87.53 \\ 
{REVEAL}     & 93.75 & \cellcolor{best}\textbf{97.81} & \cellcolor{best}\textbf{97.19}  & \cellcolor{second}\underline{95.00} & 86.88 & \cellcolor{second}\underline{96.25} & \cellcolor{second}\underline{95.94}  & \cellcolor{second}\underline{96.88} & \cellcolor{second}\underline{94.96} \\
{Veritas}     & 66.06 & 79.73 & 81.11  & 90.11 & 82.46 & 82.85 & 83.84  & 85.63 & 81.47 \\
{Ivy-xDetector}     & \cellcolor{second}\underline{95.00} & 85.54 & 86.72  & 94.44 & \cellcolor{second}\underline{94.81} & 94.38 & 94.68  & 93.52 & 92.39 \\

\textit{\textbf{ClueAegis}} & \cellcolor{best}\textbf{97.60} & 92.79 & \cellcolor{second}\underline{91.76}  & \cellcolor{best}\textbf{96.38} & \cellcolor{best}\textbf{96.56} & \cellcolor{best}\textbf{96.28} & \cellcolor{best}\textbf{96.35}  & \cellcolor{best}\textbf{97.30} & \cellcolor{best}\textbf{95.63}  \\ 

\bottomrule
\end{tabular}
}

\label{table:Genimage}

\end{center}
\vspace{-6mm}
\end{table*}

\paragraph{Generalization to AIGI-Now.} As shown in Table~\ref{tab:aigi-now}, ClueAegis demonstrates strong generalization to recently advanced generative models. It surpasses the previous state-of-the-art method by \textbf{7.88\%} in accuracy. The improvement is attributed to the heuristic-to-reasoning paradigm, which enables more robust forensic reasoning under unseen generative artifacts.

\paragraph{Performance on GenImage.} As shown in Table~\ref{table:Genimage}, we evaluate ClueAegis on the GenImage benchmark under standard synthetic image detection settings. ClueAegis consistently achieves the best overall performance compared with existing methods. The results indicate strong and stable performance across diverse synthetic image sources.

\subsection{Ablation Studies}

\paragraph{Effect of Different Backbone Models.}

\begin{figure}[htbp]
    \centering
    \includegraphics[width=\linewidth]{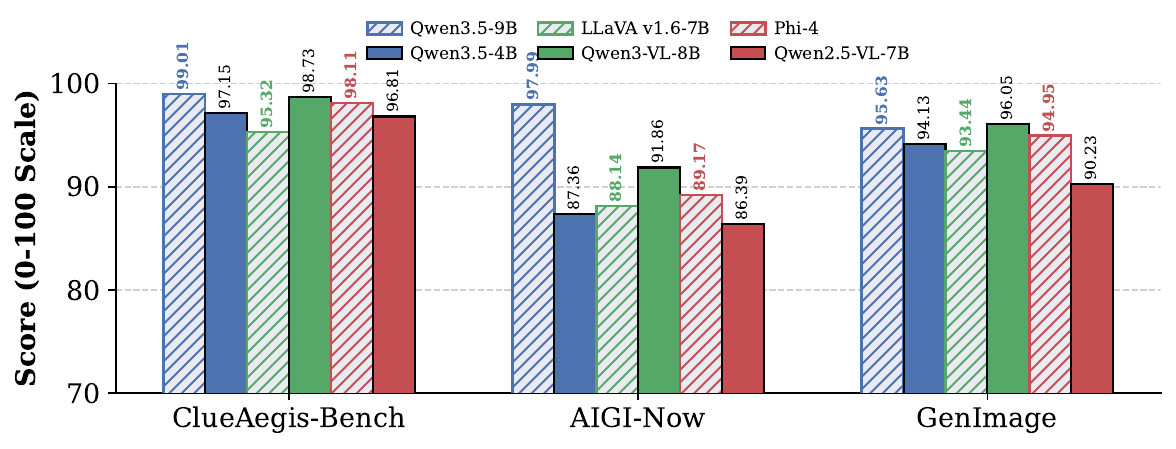}
    \caption{Performance with different backbones.}
    \label{fig:different_backbone}
    \vspace{-4mm}
\end{figure}
As shown in Figure~\ref{fig:different_backbone}, ClueAegis consistently benefits from stronger backbone models across the Qwen series, Phi-4~\cite{abouelenin2025phi}, and LLaVA-1.6~\cite{liu2023visual}. Overall, larger and more capable vision-language models achieve better detection performance, especially on challenging unseen benchmarks such as AIGI-Now. This suggests that ClueAegis relies on strong multimodal perception and effective skill activation to support forensic reasoning.

\paragraph{Effect of Forensic Skill Numbers.} As shown in Table~\ref{tab:skill_scale}, we evaluate the impact of different numbers of forensic skills by reporting the average performance across ClueAegis-Bench, AIGI-Now, and GenImage. Experimental results show that increasing the number of forensic skills improves detection performance, demonstrating that richer skill-aware cognitive routing enables the model to capture more diverse and heterogeneous forgery patterns across synthetic image distributions.

\begin{table}[t]
\centering
\small
\caption{Effect of cognitive skill scale on detection performance across different benchmarks.}
\renewcommand{\arraystretch}{1.15}
\setlength{\tabcolsep}{4pt}
\resizebox{\columnwidth}{!}{
\begin{tabular}{lccc}
\toprule
Skill Configuration & AEGIS-Bench & REVEAL-Bench &  GenImage \\
\midrule
None (0 Skills)     & 87.31 & 90.94 & 89.79 \\
Compact (4 Skills)  & 91.53 & 94.73 & 93.24 \\
Advanced (8 Skills) & 95.68 & 95.74& 94.28 \\
Full (12 Skills)    & 99.01 & 97.99 & 95.63 \\
\bottomrule
\end{tabular}
}
\label{tab:skill_scale}
\end{table}
\paragraph{Effect of Training Strategy.}

\begin{table}[t]
\centering
\small
\caption{Ablation of training components.}
\resizebox{\linewidth}{!}{
\begin{tabular}{ccccc}
\toprule
Base Model &  Skill &  SFT &  GRPO Alignment & ACC (\%) \\
\midrule
\textcolor{green}{\checkmark} & \textcolor{red}{\xmark} & \textcolor{red}{\xmark} & \textcolor{red}{\xmark} & 57.82 \\

\textcolor{green}{\checkmark} & \textcolor{green}{\checkmark} & \textcolor{red}{\xmark} & \textcolor{red}{\xmark} & 59.97 \\

\textcolor{green}{\checkmark} & \textcolor{red}{\xmark} & \textcolor{green}{\checkmark} & \textcolor{red}{\xmark} & 89.84 \\

\textcolor{green}{\checkmark} & \textcolor{green}{\checkmark} & \textcolor{green}{\checkmark} & \textcolor{red}{\xmark} & 96.40 \\

\textcolor{green}{\checkmark} & \textcolor{green}{\checkmark} & \textcolor{green}{\checkmark} & \textcolor{green}{\checkmark} & 97.54 \\
\bottomrule
\end{tabular}

}
\vspace{-4mm}
\label{tab:ablation_training}
\end{table}
As shown in Table~\ref{tab:ablation_training}, we report the average accuracy across ClueAegis-Bench, AIGI-Now, and GenImage to analyze the effectiveness of different training strategies. Specifically, we compare the performance of the base model, skill-guided inference, supervised fine-tuning (SFT), and GRPO-based cognitive alignment. Experimental results show that enabling the model to first perform cognitive perception and subsequently activate skill-specific forensic reasoning significantly improves synthetic image detection performance. Furthermore, SFT and GRPO alignment further enhance reasoning consistency and generalization capability across diverse synthetic image distributions.

\subsection{Robustness Evaluation}
We evaluate the robustness of ClueAegis on ClueAegis-Bench under common image perturbations, including JPEG compression, Gaussian blur, and resizing, which simulate real-world distribution shifts that often degrade synthetic image detection performance. As shown in Table~\ref{tab:robustness}, ClueAegis consistently outperforms all baselines across corruption types and severity levels, demonstrating strong stability under distribution shift. In contrast, existing methods suffer noticeable performance degradation, particularly under heavy blur and resizing. These results demonstrate that the skill-conditioned framework is robust to input corruption, as skill-level reasoning operates on abstract forensic cues that remain more invariant to low-level perturbations, rather than relying on fragile pixel-level patterns.

\begin{table}[t]
\centering
\small
\caption{Robustness comparison in terms of ACC under common image perturbations.}
\renewcommand{\arraystretch}{1.1}
\setlength{\tabcolsep}{3pt}
\resizebox{\columnwidth}{!}{
\begin{tabular}{lccccc}
\toprule
\multirow{2}{*}{Method} 
& \multicolumn{2}{c}{JPEG Compression} 
& \multicolumn{2}{c}{Gaussian Blur} 
& Resize \\
\cmidrule(lr){2-3} \cmidrule(lr){4-5}
& QF=85 & QF=70 & $\sigma = 1.0$ & $\sigma = 2.0$ & $\times 0.5$ \\
\midrule
UnivFD       & 85.19 & 83.62 & 86.41 & 84.75 & 86.13 \\
DDA           & 90.95& 88.22 & 91.54 & 85.24 & 92.98 \\
AIGI-Holmes      & 79.64 & 77.01 & 83.36 & 78.12 & 80.69 \\
Veritas              & 75.86 & 74.92 & 76.71 & 74.37 & 62.13 \\

\midrule
ClueAegis       & 95.43 & 92.84 & 92.84 & 86.25 & 94.87 \\
\bottomrule
\end{tabular}
}
\label{tab:robustness}
\end{table}

\subsection{Evaluation on Additional Datasets}
\label{additional}
\begin{figure}[t]
    \centering
    \includegraphics[width=\linewidth]{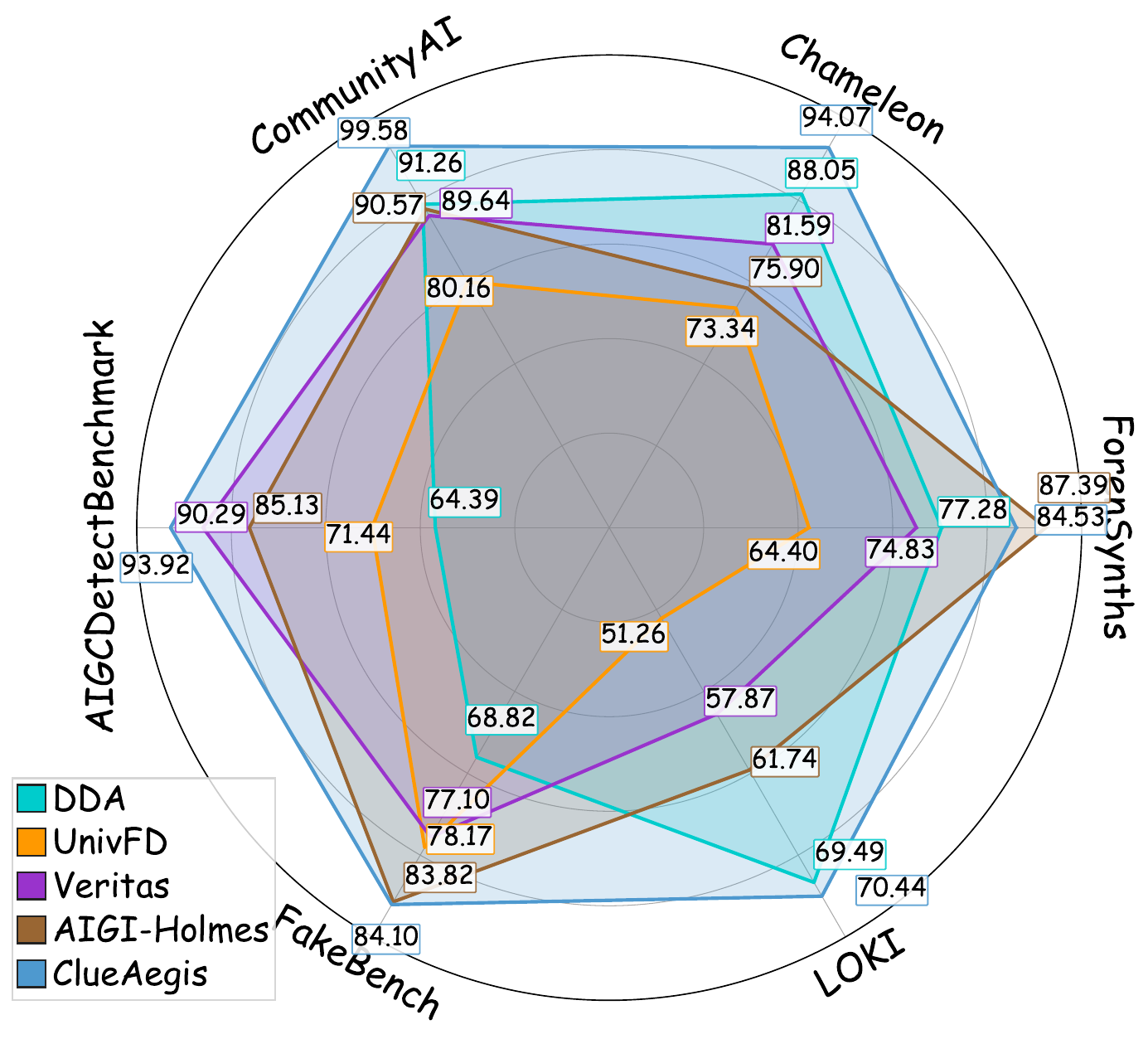}
\caption{Results on additional benchmarks.}    \label{fig:rader_additional}
\vspace{-5mm}
\end{figure}

We further evaluate ClueAegis on six additional datasets (ForenSynths~\cite{wang2020cnn}, AIGCDetectBenchmark~\cite{zhong2023patchcraft}, Chameleon~\cite{yan2024sanity}, FakeBench~\cite{li2025fakebench}, LOKI~\cite{ye2025loki}, and CommunityAI~\cite{li2026artificial}) and compare it with several state-of-the-art methods. As shown in Fig.~\ref{fig:rader_additional}, ClueAegis achieves the best or highly competitive performance across all datasets, obtaining the top results on most benchmarks while remaining comparable to the strongest baseline on ForenSynths. These results demonstrate its strong generalization ability under diverse data distributions.

\section{Conclusion}

In this paper, we revisit synthetic image detection from a heuristic-to-reasoning perspective and propose \textbf{ClueAegis}, a skill-driven forensic framework that transforms end-to-end detection into structured reasoning. By decomposing authenticity analysis into forensic skills, \textbf{ClueAegis} enables adaptive skill routing and evidence verification across diverse forgeries. We further introduce \textbf{ClueAegis-Bench}, a large-scale benchmark with explicit skill annotations for structured detection. Experiments demonstrate SOTA performance with strong generalization across diverse synthetic distributions. This work promotes skill-aware reasoning for trustworthy synthetic media detection.

\section*{Limitations}

Despite its effectiveness, ClueAegis has several limitations. First, the framework relies on a predefined set of forensic skills, which may not fully cover emerging or unseen manipulation patterns, limiting adaptability to novel generation techniques. Second, the performance of ClueAegis is partially dependent on the capability of the underlying vision-language backbone; weaker models may struggle to reliably activate and ground the correct forensic skills, leading to degraded reasoning quality. Third, the current skill annotations in ClueAegis-Bench are manually designed and may introduce bias or ambiguity in borderline cases where multiple forensic cues coexist. Finally, the structured reasoning pipeline introduces additional computational overhead compared to direct binary classification, which may affect scalability in real-world low-latency scenarios.

\section*{Ethics Statement}

This work aims to improve the reliability and robustness of synthetic image detection for trustworthy AI-generated media analysis. ClueAegis and ClueAegis-Bench are designed to support forensic reasoning research and help mitigate risks related to misinformation and manipulated visual content. However, synthetic image detection systems may still produce false predictions, especially under unseen generation distributions or ambiguous forensic cues. In addition, the predefined forensic skill annotations may introduce subjective bias in certain cases. We therefore encourage future work on more adaptive, fair, and robust reasoning mechanisms for synthetic media forensics. All datasets used in this work are collected from publicly available sources, and no private or personally sensitive information is intentionally introduced.


\bibliography{main}

\clearpage

\appendix
\nolinenumbers
\section{Visualization of ClueAegis-Bench}
\label{visualization}
Figure~\ref{fig:visualization} presents the visualization of ClueAegis-Bench. The benchmark organizes images into skill-driven forensic subspaces according to distinct cognitive reasoning cues, covering diverse synthetic image distributions and visual contents. Each subspace corresponds to a specific forensic skill and is associated with structured reasoning annotations, enabling skill-aware evaluation beyond conventional real-versus-fake classification. Specifically, \textit{Lighting Consistency} focuses on illumination-related artifacts and abnormal lighting patterns; \textit{Shadow Consistency} examines the coherence between shadows and corresponding objects; \textit{Physical Consistency} analyzes violations of physical laws and spatial plausibility; \textit{Common Sense Consistency} targets semantically unreasonable or commonsense-defying content; \textit{Functional Consistency} evaluates incorrect functional structures or tool usage; \textit{OCR Consistency} analyzes textual content and typography within images; \textit{Human Anatomy} and \textit{Animal Anatomy} focus on structural abnormalities in humans and animals, respectively; \textit{Region Analysis} investigates local visual artifacts and regional inconsistencies; \textit{Frequency Consistency} performs spectrum-domain forensic analysis; \textit{Pixel Consistency} examines low-level pixel distributions and texture continuity; and \textit{Transformation Consistency} focuses on inconsistencies introduced by post-processing operations such as rotation and color transformation.
\begin{strip}
\begin{center}
    \centering
    \includegraphics[height=0.58\textheight,keepaspectratio]{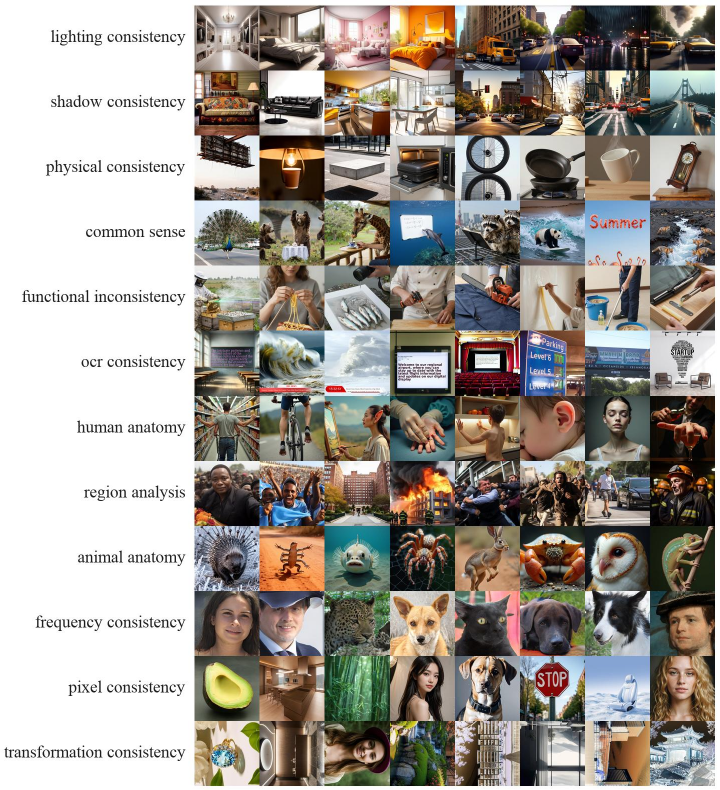}
    \captionsetup{hypcap=false}
    \captionof{figure}{\textbf{Visualization of ClueAegis-Bench.} Representative image samples from different skill-driven forensic subspaces with diverse visual contents and synthetic distributions.}
    \label{fig:visualization}
\end{center}
\end{strip}

\FloatBarrier
\clearpage

\section{Details of ClueAegis-Bench}
\label{details_bench}

\begin{figure}[t]
    \centering
    \includegraphics[width=\linewidth]{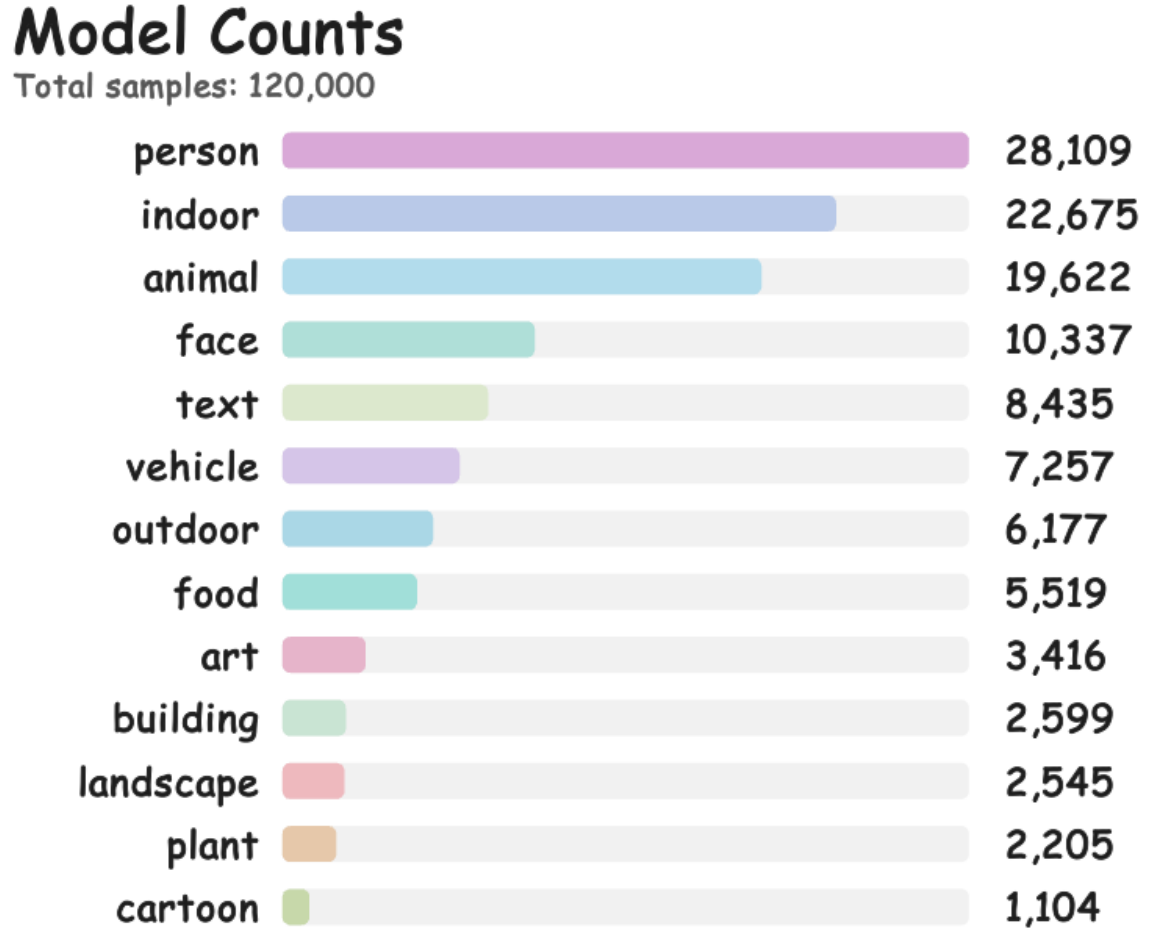}
    \caption{Performance across different visual content types.}
    \label{fig:content_type}
    
\end{figure}

As illustrated in Figure~\ref{fig:content_type}, we present the distribution of ClueAegis-Bench across different image content categories. In this section, we further describe the detailed annotation pipeline of the benchmark construction process. During data annotation, we first collected a large-scale pool of real and synthetic images from diverse sources. We then utilized the skill definitions provided in Appendix~\ref{details_skill} together with Qwen3-VL-235B to perform skill-oriented annotation. For each candidate image, the model was prompted to determine whether the image could be reliably identified through a specific forensic skill. Only images that could be correctly and consistently recognized under a single corresponding forensic skill were retained. The final benchmark therefore consists of paired \texttt{(image, skill)} annotations, where each image is explicitly associated with its dominant forensic reasoning cue. This annotation strategy enables the benchmark to establish a structured mapping between visual evidence and forensic reasoning skills, facilitating skill-aware synthetic image detection and interpretable reasoning evaluation beyond conventional binary classification settings.

\FloatBarrier

\section{Training Details}

In this section, we present the training details of our model, including the configurations for supervised fine-tuning (SFT) and Group Relative Policy Optimization (GRPO). ClueAegis-Bench contains 120K samples in total, which are split into 96K training samples and 24K held-out test samples. To ensure a fair comparison, all binary classification baseline models are trained on the same 96K training split. For large multimodal LLM-based methods, we follow their standard evaluation protocols without additional training on ClueAegis-Bench, due to their reliance on proprietary training paradigms or pretraining constraints.
\label{training_details}
\subsection{SFT Training}

\begin{figure}[t]
    \centering

    \begin{subfigure}{0.48\linewidth}
        \centering
        \includegraphics[width=\linewidth]{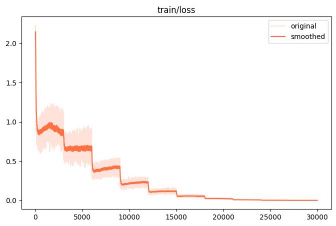}
        \caption{Train loss}
        \label{fig:train_loss}
    \end{subfigure}
    \hfill
    \begin{subfigure}{0.48\linewidth}
        \centering
        \includegraphics[width=\linewidth]{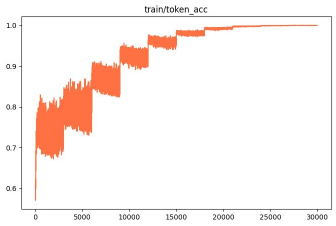}
        \caption{Train token accuracy}
        \label{fig:train_acc}
    \end{subfigure}

    \caption{Training dynamics of our model. (a) Training loss. (b) Token-level accuracy.}
    \label{fig:train_curve}
\end{figure}

We fine-tune the model using full-parameter supervised fine-tuning (SFT) on 32 NVIDIA H20 GPUs (96GB memory each) in a distributed multi-node setting. We adopt the Qwen3.5-9B model as the backbone and train with bfloat16 precision. The model is optimized using a learning rate of $1\times10^{-4}$ with a batch size of 32 (via gradient accumulation). We train for 10 epochs with a maximum sequence length of 7168 tokens. The warmup ratio is set to 0.05, and we update all linear modules during training. DeepSpeed ZeRO-2 is employed to reduce memory consumption and enable efficient large-scale training. Figure~\ref{fig:train_curve} illustrates the training dynamics, including the convergence behavior of the training loss and token-level accuracy.

\FloatBarrier

\subsection{GRPO Training}
We further fine-tune the model using Group Relative Policy Optimization (GRPO) on top of the supervised checkpoint. Training is conducted on 8 NVIDIA H20 GPUs (96GB memory each) with bfloat16 precision. We adopt a two-turn interaction setting with a maximum sequence length of 10240 tokens. The model is optimized using a cosine learning rate schedule with an initial learning rate of $1\times10^{-6}$ and trained for 1 epoch. The effective batch size is 32 via gradient accumulation. We employ a group-based reward strategy with multiple signals, including skill consistency, reasoning validity, and answer correctness, to guide policy optimization, where all reward weights are set to 1 without additional hyperparameter tuning. DeepSpeed ZeRO-2 is used for memory-efficient training.
\section{F1-score Evaluation Results}
\label{F1}

In the main paper, we report accuracy (ACC) as the primary metric under a unified real/fake decision rule. To provide a complementary view, Appendix~\ref{F1} reports F1 scores calculated from the same binary predictions. For each evaluation slice, F1 is calculated as the harmonic mean of precision and recall:
\begin{equation}
    \mathrm{F1} = \frac{2 \cdot \mathrm{Precision} \cdot \mathrm{Recall}}{\mathrm{Precision} + \mathrm{Recall}}.
\end{equation}
We calculate F1 separately for each forensic skill on ClueAegis-Bench, each generator--cue pair on AIGI-Now, and each generator family on GenImage. The average column follows the aggregation protocol of the corresponding benchmark. These results provide a class-sensitive view of detector performance, especially on slices where class imbalance or uneven difficulty can make accuracy less informative.

\subsection{F1-score Evaluation on ClueAegis-Bench}
\label{app:f1_clueaegis_bench}

Table~\ref{tab:AEGIS-Bench-F1} mirrors Table~\ref{tab:AEGIS-Bench} by replacing accuracy with F1 on the twelve skill-specific test subsets. ClueAegis achieves the strongest average F1 score (\textbf{99.00\%}) and ranks first on every skill, indicating consistently strong class-sensitive performance across diverse forensic dimensions. Among prior methods, DDA obtains the highest average F1 score (\underline{91.25\%}), followed by AIDE (\underline{89.12\%}). Several specialized detectors remain competitive on isolated skills, such as CNNSpot on Region and AIGI-Holmes on Freq/Pixel, but their performance varies substantially across other skills. Overall, the F1 results support the accuracy-based conclusion that cognition-guided skill routing is most beneficial when single-feature detectors become brittle.

\begin{table*}[htbp]
    \centering
    \caption{\textbf{F1-score Evaluation on ClueAegis-Bench.} F1 is computed for each skill-specific test subset with \textit{Fake} as the positive class.}

    \label{tab:AEGIS-Bench-F1}
    \footnotesize
    \renewcommand\arraystretch{1.15}

    \resizebox{\textwidth}{!}{
    \begin{tabular}{c ccccccccccccc c}
        \toprule
        \multirow{2}{*}{\textbf{Method}} 
        & \multicolumn{12}{c}{\textbf{Forensic Consistency Metrics}} 
        & \multirow{2}{*}{\textbf{Avg.}} \\
        
        \cmidrule(lr){2-13}
        
        & \textbf{Light} 
        & \textbf{Shadow} 
        & \textbf{Phys} 
        & \textbf{CS} 
        & \textbf{Func} 
        & \textbf{OCR} 
        & \textbf{Human} 
        & \textbf{Region} 
        & \textbf{Animal} 
        & \textbf{Freq} 
        & \textbf{Pixel} 
        & \textbf{Trans} \\
        
        \midrule
        CNNSpot & 94.35 & 93.29 & 91.70 & 92.39 & 91.80 & 92.20 & 92.72 & \cellcolor{second}\underline{87.56} & 92.50 & 65.35 & 91.33 & 82.07 & 88.94 \\
        FreqNet & 84.43 & 84.75 & 89.79 & 93.68 & 92.50 & 87.06 & 92.37 & 32.73 & 90.49 & 86.55 & 88.75 &75.03 & 83.18 \\
        UnivFD & 93.49 & 91.37 & 86.95 & 94.28 & 90.39 & 90.39 & 91.68 & 71.73 & 92.68 & 81.67 & 82.17 & 75.56 & 86.86 \\
        NPR & 87.21 & 82.42 & 88.37 & 89.16 & 88.90 & 85.21 & 88.67 & 82.10 & 88.53& 89.12 & 86.04 & 84.68 & 86.70 \\
        HyperDet & 1.19 & 0.40 & 84.85 & 88.65 & 84.49 & 1.78 & 89.67 & 0.80 & 89.21 & 84.68 & 75.39 & 54.47 & 54.63 \\
        AIDE & \cellcolor{second}\underline{98.89} & 90.62 & 94.20 & 94.50 & 94.20 & \cellcolor{second}\underline{93.18} & 94.50 & 45.92 & 94.25 & 93.50 & 92.33 & 83.38 & 89.12 \\
        DDA & 94.50 & \cellcolor{second}\underline{93.70} & \cellcolor{second}\underline{94.45} & 94.65 & \cellcolor{second}\underline{94.25} & 93.10 & 94.50 & 63.30 & 94.50 & 94.35 & 93.58 & \cellcolor{second}\underline{90.18} & \cellcolor{second}\underline{91.25} \\
        AIGI-Holmes & 67.45 & 67.14 & 86.18 & 89.90 & 80.49 & 90.73 & 90.75 & 82.72 & 80.68 & \cellcolor{second}\underline{98.96} & \cellcolor{second}\underline{97.76} & 84.69 & 84.79 \\
        Veritas & 87.03 & 79.12 & 71.69 & 79.20 & 70.85 & 51.57 & 85.68 & 68.39 & 77.02 & 76.55 & 70.64 & 64.32 & 73.50 \\
        Ivy-xDetector & 66.96 & 67.07 & 93.31 & \cellcolor{second}\underline{95.00} & 91.73 & 63.52 & \cellcolor{second}\underline{94.93} & 81.68 & \cellcolor{second}\underline{95.16} & 76.65 & 90.33 & 73.16 & 82.46 \\
        \textit{\textbf{ClueAegis}} & \cellcolor{best}\textbf{99.80} & \cellcolor{best}\textbf{99.70} & \cellcolor{best}\textbf{99.45} & \cellcolor{best}\textbf{99.70} & \cellcolor{best}\textbf{99.75} & \cellcolor{best}\textbf{99.85} & \cellcolor{best}\textbf{99.70} & \cellcolor{best}\textbf{95.49} & \cellcolor{best}\textbf{99.75} & \cellcolor{best}\textbf{99.95} & \cellcolor{best}\textbf{99.50} & \cellcolor{best}\textbf{95.35} & \cellcolor{best}\textbf{99.00} \\
        \bottomrule
    \end{tabular}
    }
\end{table*}

\subsection{F1-score Evaluation on AIGI-Now}
\label{app:f1_aigi_now}

Table~\ref{tab:aigi-now-F1} reports F1 scores on AIGI-Now across recent generators and two cue types, \textit{pix} and \textit{sem}. ClueAegis achieves the best average F1 score (\textbf{97.93\%}), outperforming the strongest prior baseline, Ivy-xDetector (\underline{91.51\%}), by 6.42\% points. The improvement is especially consistent on semantic-cue subsets, where ClueAegis obtains near-saturated F1 scores across all generators. These results suggest that the proposed heuristic-to-reasoning pipeline generalizes beyond low-level artifacts and remains effective on challenging semantic manipulations.
\begin{table*}[htbp]
\centering
\caption{\textbf{F1-score Evaluation on AIGI-Now.} F1 is computed for each generator--cue pair with \textit{Fake} as the positive class.}
\label{tab:aigi-now-F1}
\small
\resizebox{\textwidth}{!}{
\begin{tabular}{l|cc|cc|cc|cc|cc|cc|cc|cc|cc|c}
\toprule
\multirow{2}{*}{\textbf{Detector}} & 
\multicolumn{2}{c|}{\textbf{FLUX-dev}} & 
\multicolumn{2}{c|}{\textbf{FLUX-kera}} & 
\multicolumn{2}{c|}{\textbf{FLUX-kontext}} & 
\multicolumn{2}{c|}{\textbf{FLUX-pro}} & 
\multicolumn{2}{c|}{\textbf{gpt4o}} & 
\multicolumn{2}{c|}{\textbf{jimeng}} & 
\multicolumn{2}{c|}{\textbf{keling}} & 
\multicolumn{2}{c|}{\textbf{minimax}} & 
\multicolumn{2}{c|}{\textbf{Nano}} & 
\multirow{2}{*}{\textbf{Avg.}} \\
& pix & sem & pix & sem & pix & sem & pix & sem & pix & sem & pix & sem & pix & sem & pix & sem & pix & sem & \\

\midrule
\multicolumn{20}{c}{\textbf{Conventional Binary Detection Methods}} \\
CNNSpot & 81.03 & 66.49 & 82.48 & 30.77 & 65.21 & 26.27 & 69.04 & 43.72 & 79.68 & 36.17 & 73.44 & 35.33 & 79.76 & 61.99 & 77.21 & 26.79 & 80.59 & 58.72  &  59.71\\
FreqNet &77.39 & 12.60 & 76.75 & 10.32 & 61.96 & 16.05 & 48.35 & 7.28 & 79.56 & 23.67 & 58.73 & 14.83 & 78.71 & 18.47 & 73.36 & 15.11 & 80.43 & 5.84&  42.19\\

UnivFD & 89.80 & \cellcolor{second}\underline{94.53} & 80.30 & \cellcolor{second}\underline{89.53} & 54.58 & 72.35 & 55.21 & \cellcolor{second}\underline{92.07} & 91.06 & \cellcolor{second}\underline{96.79} & 69.95 & \cellcolor{second}\underline{96.70} & 89.78 & \cellcolor{second}\underline{95.45} & 66.50 & 83.72 & 88.60 & \cellcolor{second}\underline{97.39}  &  83.57\\

NPR & 68.83 & 42.44 & 69.37 & 37.05 & 66.71 & 36.09 & 63.30 & 43.92 & 68.59 & 45.88 & 65.83 & 37.48 & 68.59 & 65.36 & 68.25 & 37.99 & 69.04 & 51.59  &  55.91\\

AIDE & 88.86 & 62.28 & 87.52 & 39.14 & 75.71 & 37.60 & 69.85 & 58.61 & 87.54 & 64.10 & 69.71 & 59.97 & 88.34 & 74.03 & 82.51 & 43.27 & 89.76 & 84.60 &  70.19\\
DDA & 82.07 & 88.74 & 82.21 & 49.33 & 77.24 & 46.13 & 65.71 & 84.32 & 81.97 & 84.27 & 62.84 & 77.66 & 82.13 & 85.81 & 79.87 & 49.96 & 82.95 & 89.19  &  75.13\\
\midrule
\multicolumn{20}{c}{\textbf{MLLM-based Detection Methods}} \\
AIGI-Holmes &\cellcolor{second}\underline{98.68} & 82.61 & 89.63 & 66.62 & 87.01 & 65.82 & \cellcolor{best}\textbf{98.89} & 76.09 & \cellcolor{best}\textbf{99.55} & 75.02 & \cellcolor{best}\textbf{99.75} & 82.24 & \cellcolor{second}\underline{97.34} & 85.21 & 86.46 & 49.55 & \cellcolor{second}\underline{96.38} & 90.01&  84.83\\

Veritas& 84.13 & 83.16 & 60.63 & 76.21 & 60.73 & 74.77 & 47.49 & 79.36 & 63.38 & 82.52 & 60.95 & 83.29 & 84.39 & 84.63 & 45.96 & 74.71 & 80.90 & 85.31  &  72.92\\
Ivy-xDetector & 95.59 & 89.51 & \cellcolor{second}\underline{94.94} & 88.17 & \cellcolor{second}\underline{92.23} & \cellcolor{second}\underline{82.70} & \cellcolor{second}\underline{93.79} & 88.66 & 95.62 & 88.92 & \cellcolor{second}\underline{95.80} & 89.19 & 95.72 & 89.03 & \cellcolor{second}\underline{94.85} & \cellcolor{second}\underline{88.09} & 95.51 & 88.92  &  \cellcolor{second}\underline{91.51}\\
\textit{\textbf{ClueAegis}} &\cellcolor{best}\textbf{98.71} & \cellcolor{best}\textbf{99.45} & \cellcolor{best}\textbf{98.72} & \cellcolor{best}\textbf{99.45} & \cellcolor{best}\textbf{96.18} & \cellcolor{best}\textbf{99.65} & 89.98 & \cellcolor{best}\textbf{99.21} & \cellcolor{second}\underline{98.42} & \cellcolor{best}\textbf{99.21} & 91.65 & \cellcolor{best}\textbf{99.21} & \cellcolor{best}\textbf{98.38} & \cellcolor{best}\textbf{99.16} & \cellcolor{best}\textbf{97.92} & \cellcolor{best}\textbf{99.45} & \cellcolor{best}\textbf{98.33} & \cellcolor{best}\textbf{99.65}  &  \cellcolor{best}\textbf{97.93}\\

\bottomrule
\end{tabular}
}
\end{table*}

\subsection{F1-score Evaluation on GenImage}
\label{app:f1_genimage}

Table~\ref{table:Genimage-F1} presents F1 scores on the GenImage benchmark. ClueAegis achieves the best mean F1 score (\textbf{93.92\%}), exceeding Ivy-xDetector (\underline{92.35\%}) by 1.57\% points and obtaining the top score on seven of the eight generator families. The only exception is Wukong, where Ivy-xDetector performs best while ClueAegis remains second. This trend shows that ClueAegis retains strong F1 performance under conventional synthetic-image detection settings, including generator families that differ substantially from the skill-oriented training distribution.

\begin{table*}[htbp]
    \centering
    \caption{\textbf{F1-score Evaluation on GenImage.} F1 is computed for each generator family with \textit{Fake} as the positive class.}
    \label{table:Genimage-F1}
    \footnotesize
    \renewcommand\arraystretch{1.15}
    \resizebox{\textwidth}{!}{
    \begin{tabular}{lccccccccc}
        \toprule
        \textbf{Method} & \textbf{Midjourney} & \textbf{SD v1.4} & \textbf{SD v1.5} & \textbf{ADM} & \textbf{GLIDE} & \textbf{Wukong} & \textbf{VQDM} & \textbf{BigGAN} & \textbf{Avg.} \\
        \midrule
        CNNSpot & 77.71 & 72.79 & 71.87 & 59.23 & 69.55 & 67.95 & 42.43 & 33.84 & 61.92 \\
        FreqNet & 65.06 & 59.35 & 57.59 & 17.71 & 53.08 & 57.70 & 42.64 & 32.88 & 48.25 \\
        UnivFD & 75.70 & 74.95 & 72.53 & 49.66 & 58.25 & 72.27 & 58.61 & 48.64 & 63.83 \\
        NPR & 71.30 & 71.55 & 69.77 & 71.90 & 69.82 & 70.69 & 68.58 & 60.49 & 69.26 \\
        AIDE & 82.02 & 78.70 & 79.00 & 71.81 & 83.42 & 78.14 & 72.56 & 58.79 & 75.55 \\
        DDA & 83.34 & 83.08 & 82.23 & 78.29 & 79.43 & 82.00 & 82.74 & 83.03 & 81.77 \\
        Veritas & 50.18 & 61.87 & 63.56 & 61.60 & 63.54 & 71.53 & 69.14 & 66.33 & 63.47 \\
        Ivy-xDetector & \cellcolor{second}\underline{94.25} & \cellcolor{second}\underline{85.71} & \cellcolor{second}\underline{87.44} & \cellcolor{second}\underline{93.47} & \cellcolor{second}\underline{95.14} & \cellcolor{best}\textbf{95.48} & \cellcolor{second}\underline{94.64} & \cellcolor{second}\underline{92.64} & \cellcolor{second}\underline{92.35} \\
        \textit{\textbf{ClueAegis}} & \cellcolor{best}\textbf{96.33} & \cellcolor{best}\textbf{90.31} & \cellcolor{best}\textbf{90.48} & \cellcolor{best}\textbf{95.27} & \cellcolor{best}\textbf{96.48} & \cellcolor{second}\underline{90.59} & \cellcolor{best}\textbf{96.06} & \cellcolor{best}\textbf{95.87} & \cellcolor{best}\textbf{93.92} \\
        \bottomrule
    \end{tabular}
    }
\end{table*}

\section{Few-Shot Performance Evaluation}
\label{app:few_shot}

To further examine data efficiency, we evaluate all methods under a few-shot setting on ClueAegis-Bench using the same skill-wise accuracy protocol as the main evaluation. As shown in Table~\ref{tab:few-shot}, ClueAegis achieves the best average accuracy (\textbf{97.60\%}) and ranks first across all twelve forensic skills. Compared with the strongest baseline, DDA (\underline{92.42\%}), ClueAegis improves the average accuracy by 5.18\% points. The gains are especially pronounced on OCR, Region, Pixel, and Trans, where reliable detection requires reasoning over localized evidence or structural transformations rather than relying only on global visual artifacts. These results indicate that the proposed heuristic-to-reasoning framework remains effective when only limited supervision is available, suggesting stronger data efficiency and skill-level generalization than conventional detectors.

\begin{table*}[htbp]
    \centering
    \caption{\textbf{Few-Shot Performance Evaluation.} Accuracy is reported on ClueAegis-Bench under the few-shot setting.
    }

    \label{tab:few-shot}
    \footnotesize
    \renewcommand\arraystretch{1.15}

    \resizebox{\textwidth}{!}{
    \begin{tabular}{c ccccccccccccc c}
        \toprule
        \multirow{2}{*}{\textbf{Method}} 
        & \multicolumn{12}{c}{\textbf{Forensic Consistency Metrics}} 
        & \multirow{2}{*}{\textbf{Avg.}} \\
        
        \cmidrule(lr){2-13}
        
        & \textbf{Light} 
        & \textbf{Shadow} 
        & \textbf{Phys} 
        & \textbf{CS} 
        & \textbf{Func} 
        & \textbf{OCR} 
        & \textbf{Human} 
        & \textbf{Region} 
        & \textbf{Animal} 
        & \textbf{Freq} 
        & \textbf{Pixel} 
        & \textbf{Trans} \\
        
        \midrule
        CNNSpot & 91.60 & 92.10 & 90.10 & 90.70 & 91.05 & \cellcolor{second}\underline{92.30} & 91.15 & 87.80 & 90.50 & 89.35 & 85.79 & 75.15 & 88.97 \\
        FreqNet & 72.90 & 70.50 & 86.10 & 91.20 & 90.65 & 83.50 & 92.60 & 77.95 & 91.35 & 79.60 & 81.73 & 71.55& 82.47 \\
        UnivFD & \cellcolor{second}\underline{93.50} & 91.50 & 87.05 & 90.55 & 89.30 & 85.70 & 91.70 & 71.95 & 90.10 & 76.50 & 77.23 & 70.90 & 84.67 \\
        NPR & 85.25 & 75.30 & 90.05 & 92.55 & 91.80 & 67.85 & 92.95 & 78.00 & 91.00 & 90.60 & 84.69 & 71.25 & 84.27 \\
        HyperDet & 59.15 & 49.20 & 91.20 & 94.10 & 93.00 & 58.25 & 92.25 & 69.90 & 90.05 & 87.85 & 83.89 & 75.80 & 78.72 \\
        AIDE & 93.30 & 91.90 & 91.35 & 92.65 & 91.45 & 88.50 & 93.15 & 76.15 & 92.25 & 91.15 & 88.29 & 77.85 & 89.00 \\
        DDA & 92.65 & \cellcolor{second}\underline{93.90} & \cellcolor{second}\underline{93.90} & \cellcolor{second}\underline{94.40} & \cellcolor{second}\underline{93.90} & 91.35 & \cellcolor{second}\underline{94.20} & \cellcolor{second}\underline{88.10} & \cellcolor{second}\underline{94.30} & \cellcolor{second}\underline{93.90} & \cellcolor{second}\underline{92.40} & \cellcolor{second}\underline{86.00} & \cellcolor{second}\underline{92.42} \\

        \textit{\textbf{ClueAegis}} & \cellcolor{best}\textbf{98.55} & \cellcolor{best}\textbf{99.00} & \cellcolor{best}\textbf{97.40} & \cellcolor{best}\textbf{98.35} & \cellcolor{best}\textbf{97.75} & \cellcolor{best}\textbf{99.55} & \cellcolor{best}\textbf{96.70} & \cellcolor{best}\textbf{94.35} & \cellcolor{best}\textbf{96.55} & \cellcolor{best}\textbf{98.90} & \cellcolor{best}\textbf{98.85} & \cellcolor{best}\textbf{95.30} & \cellcolor{best}\textbf{97.60} \\
        \bottomrule
    \end{tabular}
    }
\end{table*}

\FloatBarrier

\nolinenumbers
\section{Skill Design Details}
\label{details_skill}

This appendix reports the skill-conditioned second-round prompt templates used by ClueAegis. To avoid repeating identical instructions, we factor out the shared wrapper: each prompt is applied after the first-round image analysis, requires the model to compare the current evidence with the previous-round original image, and follows the same answer format, where the reasoning process is generated within the \texttt{<think>} block, and the final answer outputs \texttt{0} if the image is real and \texttt{1} if the image is synthetic.

Due to space limitations, we only present the detailed prompt design of the \textit{Lighting Consistency} skill in this appendix, while the complete skill-specific prompt templates and implementations will be released in future public resources. Different forensic skills are associated with different auxiliary analysis modules and external evidence extraction tools. Specifically, \textit{Lighting Consistency} incorporates an external lighting analysis model\cite{pautrat2023deeplsd} for illumination reasoning; \textit{Shadow Consistency} utilizes external object and shadow extraction models\cite{wang2020instance} to analyze shadow-object coherence; \textit{OCR Consistency} employs an OCR model\cite{cui2025paddleocr30technicalreport} to extract textual information; \textit{Region Analysis} integrates a regional perception model\cite{kirillov2023segment} to capture localized object evidence; \textit{Frequency Consistency} incorporates a frequency-spectrum analysis model\cite{tan2024frequency} for spectral forensic detection; and \textit{Pixel Consistency} leverages a pixel-level forensic model\cite{tan2024rethinking} to identify low-level pixel artifacts and local inconsistency patterns.

\definecolor{promptbg}{RGB}{252,252,252}
\definecolor{promptframe}{RGB}{191,203,217}
\definecolor{prompttitle}{RGB}{236,241,247}
\definecolor{prompttext}{RGB}{30,35,42}
\lstdefinestyle{promptstyle}{
    basicstyle=\ttfamily\footnotesize\color{prompttext},
    breaklines=true,
    breakatwhitespace=true,
    columns=fullflexible,
    keepspaces=true,
    showstringspaces=false,
    frame=none,
    aboveskip=0pt,
    belowskip=0pt
}

\newlength{\PromptBoxHeight}
\newlength{\PromptBoxTallHeight}
\setlength{\PromptBoxHeight}{0.470\textheight}
\setlength{\PromptBoxTallHeight}{0.493\textheight}

\newcommand{\PromptBoxWithHeight}[3]{%
    \begin{tcolorbox}[
        height=#1,
        valign=top,
        colback=promptbg,
        colframe=promptframe,
        colbacktitle=prompttitle,
        coltitle=black,
        title={#2},
        fonttitle=\bfseries\footnotesize,
        boxrule=0.35pt,
        arc=1mm,
        left=1.4mm,
        right=1.4mm,
        top=0.8mm,
        bottom=0.8mm,
        toptitle=0.6mm,
        bottomtitle=0.6mm,
        boxsep=0pt,
        before skip=0pt,
        after skip=0pt
    ]
    \lstinputlisting[style=promptstyle]{#3}
    \end{tcolorbox}
    \vspace{1mm}
}

\newcommand{\PromptBox}[2]{%
    \PromptBoxWithHeight{\PromptBoxHeight}{#1}{#2}
}

\newcommand{\PromptBoxTall}[2]{%
    \PromptBoxWithHeight{\PromptBoxTallHeight}{#1}{#2}
}

\raggedbottom

\begin{figure}[t]
\centering

\PromptBox{Skill 1: Lighting Consistency}{1_1_lighting.txt}

\caption{Prompt template for Lighting Consistency. The prompt includes input images, auxiliary DeepLSD-based lighting analysis, and a structured reasoning checklist for real-vs-synthetic classification.}
\label{fig:lighting_prompt}

\end{figure}



\clearpage
\flushbottom


\section{Qualitative Results}
\label{Qualitative}
In this section, we present qualitative examples of ClueAegis under all 12 predefined forensic skills. Specifically, we visualize the corresponding reasoning results on both synthesized and real-world images to demonstrate how different forensic skills respond to diverse visual patterns and forensic cues. For each skill category, representative synthetic and real image examples are provided together with the corresponding forensic reasoning process and evidence analysis generated by the model. The qualitative results further illustrate the effectiveness of the proposed skill-driven reasoning framework across heterogeneous image distributions.

\begin{figure*}[!t]
    \centering
    \includegraphics[width=\textwidth]{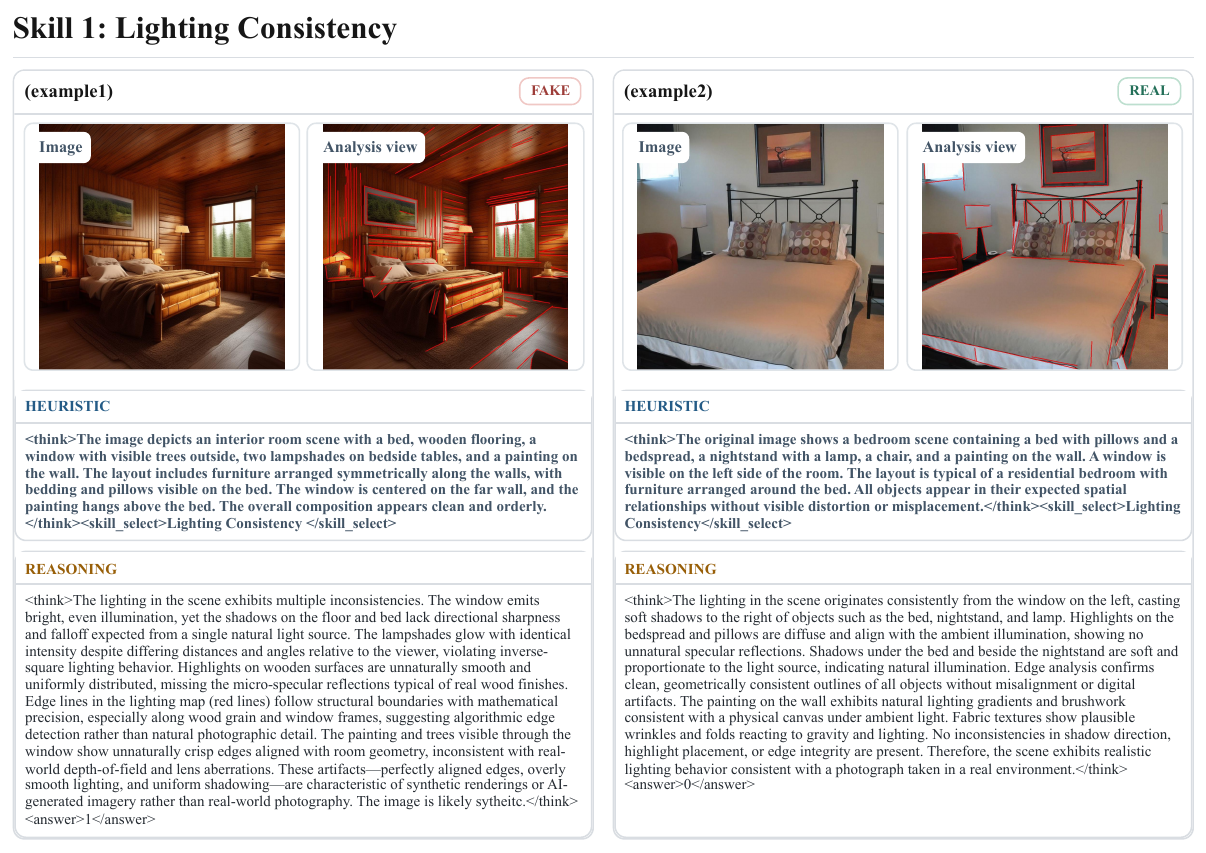}
    \caption{\textbf{Lighting Consistency.} Skill-based forensic reasoning based on illumination consistency analysis. External lighting analysis models are further incorporated to assist the examination of spatial light directions and shadow relationships.}
    \label{fig:skill_1}
\end{figure*}
\clearpage

\begin{figure*}[!t]
    \centering
    \includegraphics[width=\textwidth]{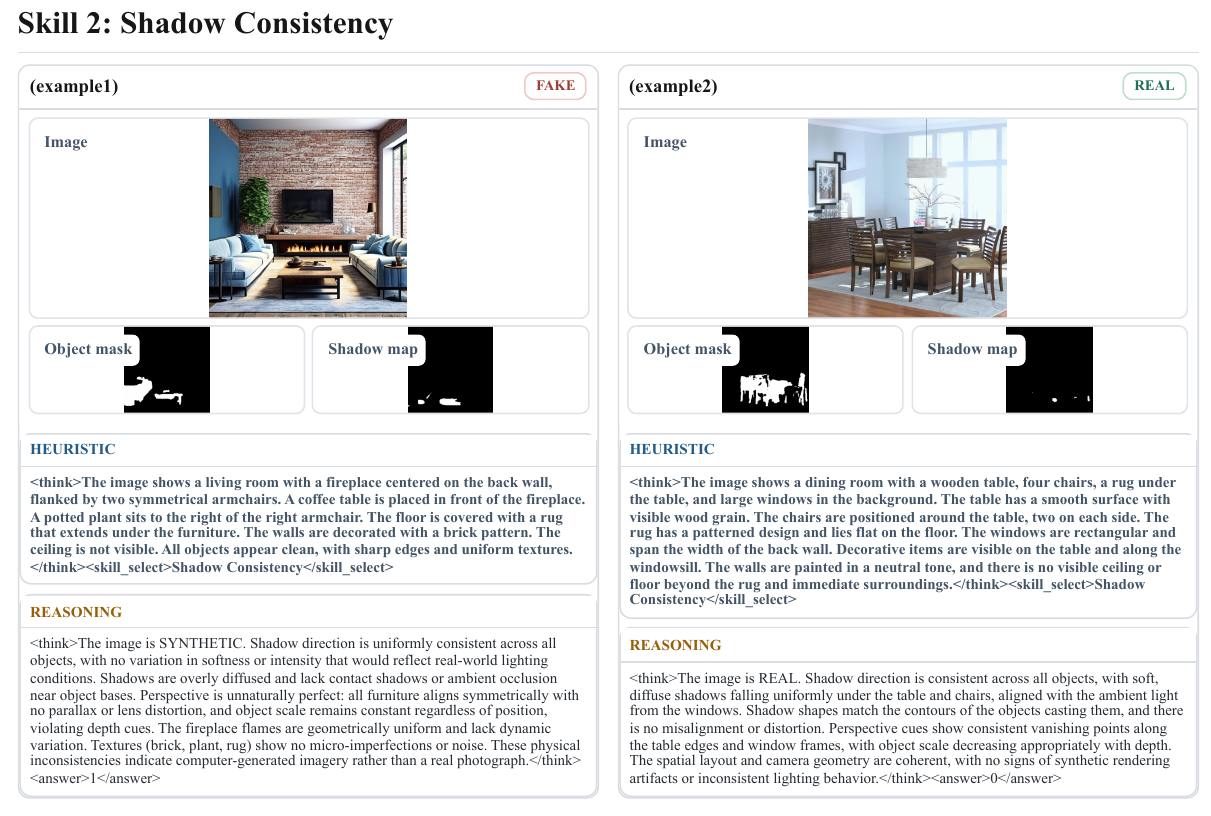}
    \caption{\textbf{Shadow Consistency.} Skill-based forensic reasoning based on shadow consistency analysis. External shadow extraction and object segmentation models are further incorporated to examine the spatial coherence among objects, illumination, and projected shadows.}
    \label{fig:skill_2}
  
\end{figure*}
\clearpage

\begin{figure*}[!t]
    \centering
    \includegraphics[width=\textwidth]{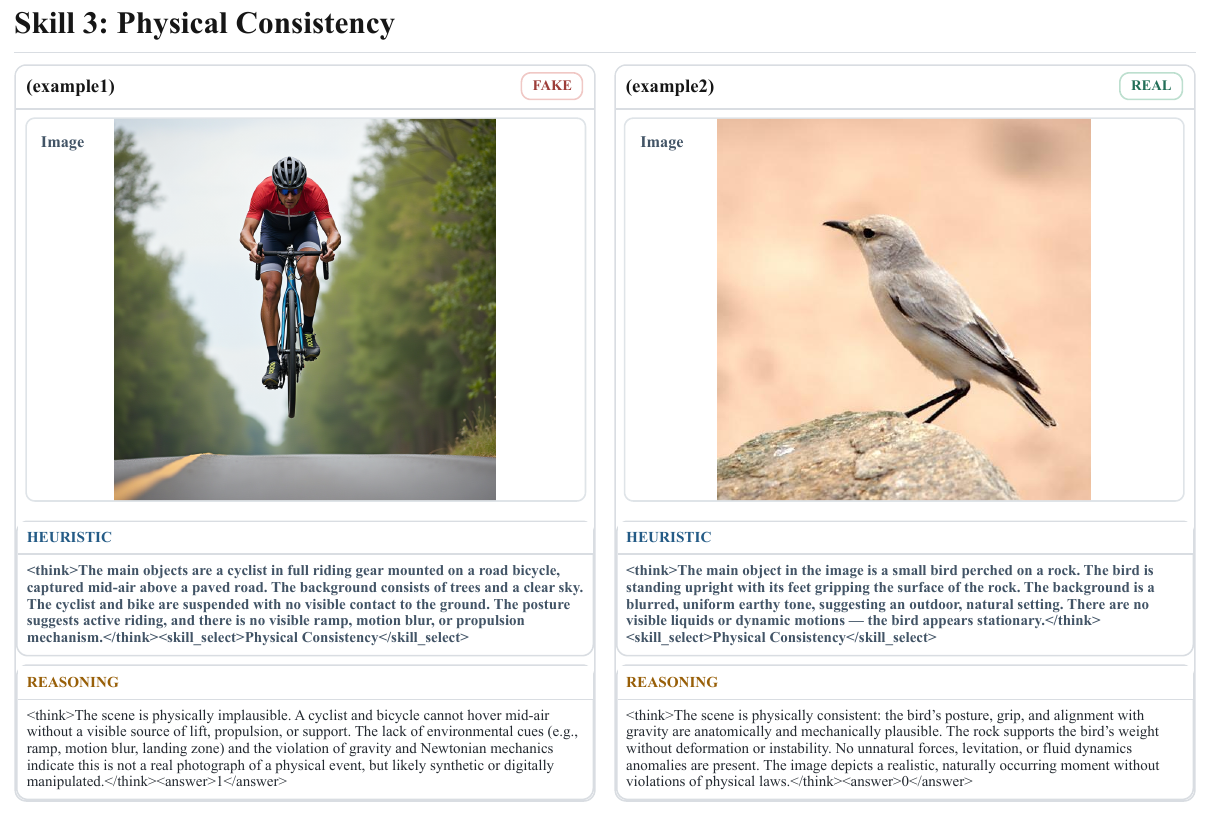}
    \caption{\textbf{Physical Consistency.} Skill-based forensic reasoning from a physical perspective, analyzing whether image content conforms to real-world physical laws and spatial interaction constraints.}
    \label{fig:skill_3}

\end{figure*}\clearpage

\begin{figure*}[!t]
    \centering
    \includegraphics[width=\textwidth]{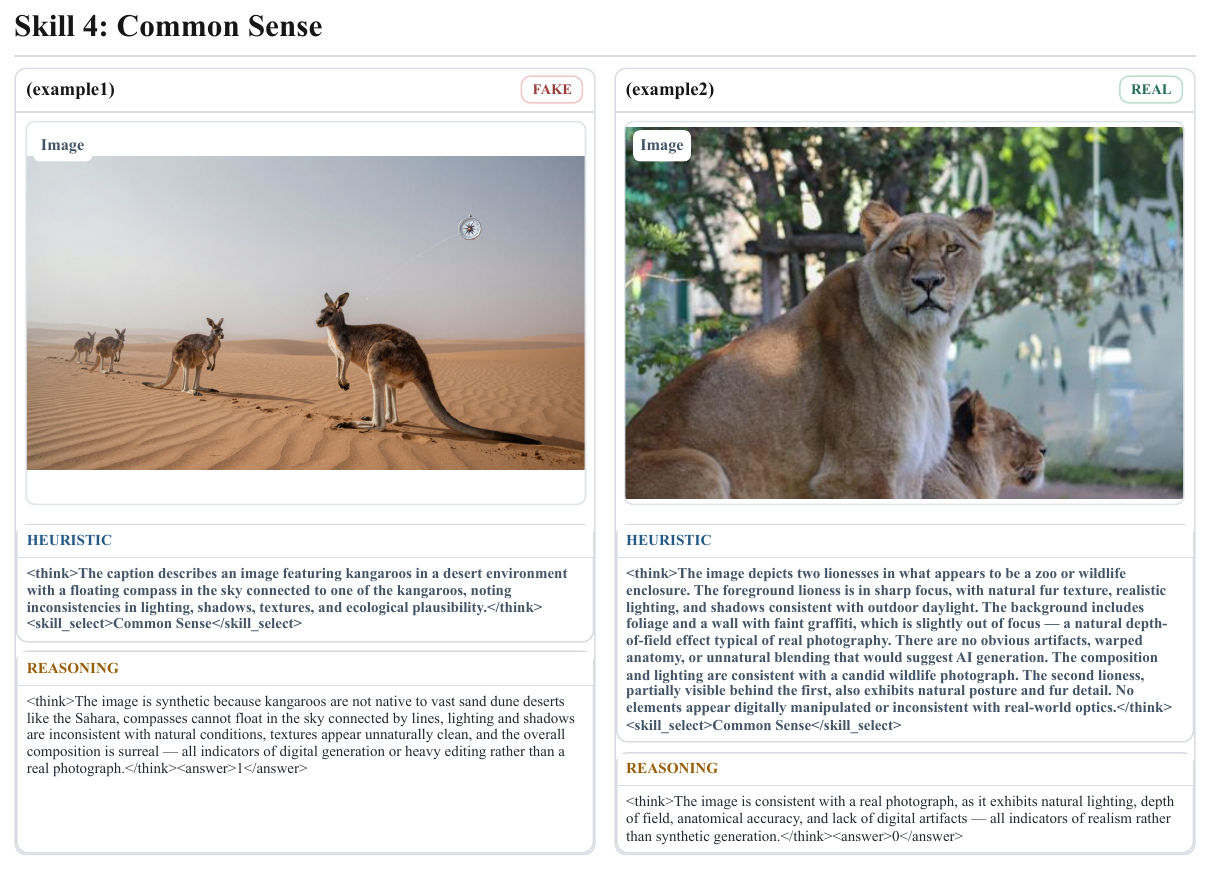}
    \caption{\textbf{Common Sense.} Skill-based forensic reasoning based on commonsense consistency analysis, evaluating whether scene content, object relationships, and semantic interactions conform to real-world expectations and everyday human cognition.}
    \label{fig:skill_4}

\end{figure*}
\clearpage

\begin{figure*}[!t]
    \centering
    \includegraphics[width=\textwidth]{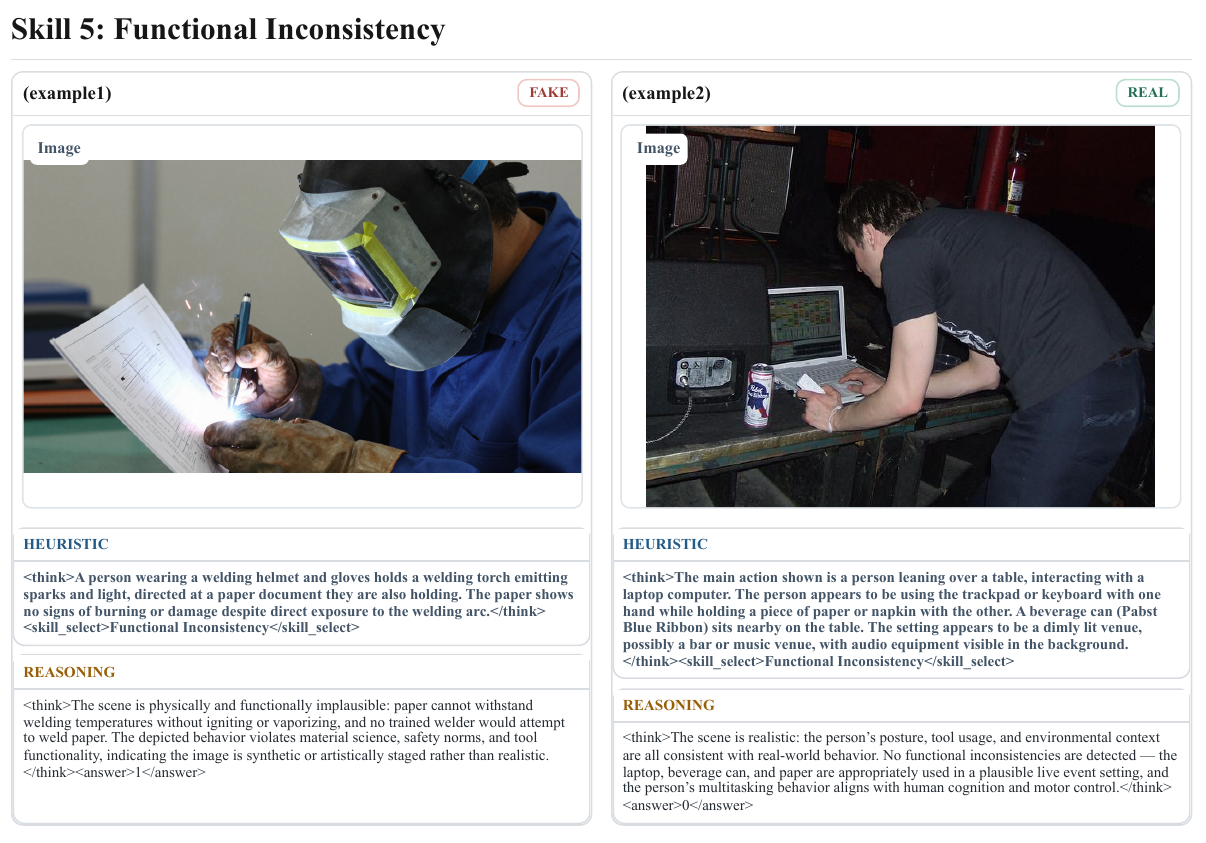}
    \caption{\textbf{Functional Inconsistency.} Skill-based forensic reasoning based on functional consistency analysis, evaluating whether objects and scene components conform to realistic functionalities and real-world usage relationships.}
    \label{fig:skill_5}

\end{figure*}

\clearpage

\begin{figure*}[!t]
    \centering
    \includegraphics[width=\textwidth]{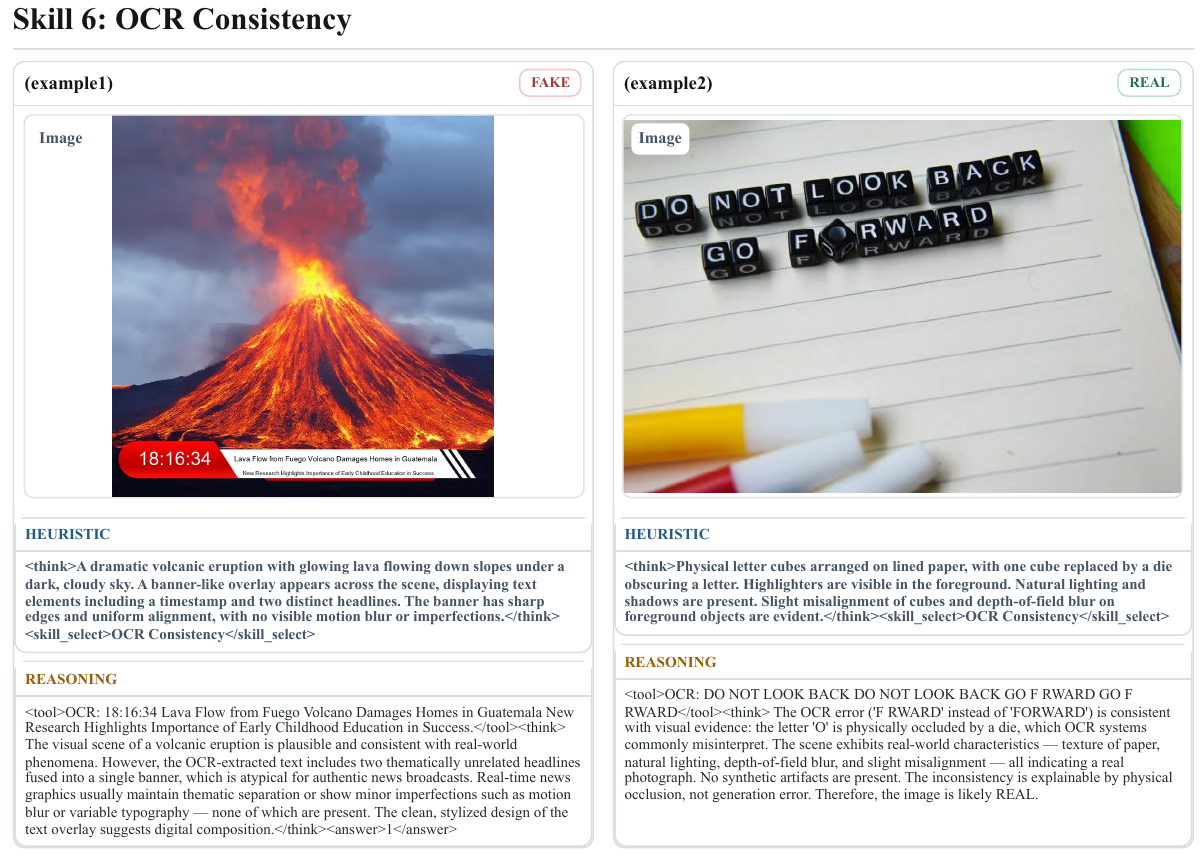}
    \caption{\textbf{OCR Consistency.} Skill-based forensic reasoning based on textual consistency analysis. External OCR models are further incorporated to examine recognized text content, typography, and semantic alignment consistency.}
    \label{fig:skill_6}
    \vspace{-4mm}
\end{figure*}
\clearpage

\begin{figure*}[!t]
    \centering
    \includegraphics[width=\textwidth]{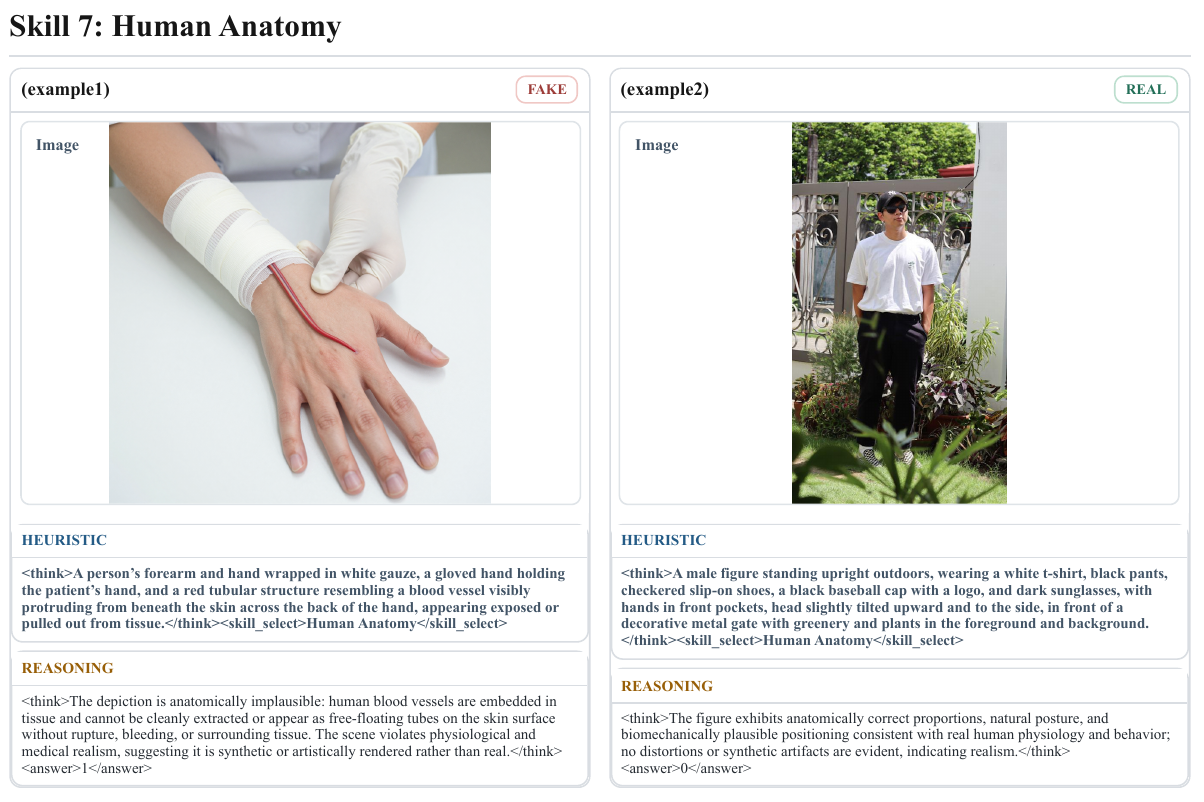}
    \caption{\textbf{Human Anatomy.} Skill-based forensic analysis that evaluates whether human body structures, anatomical proportions, and articulation relationships conform to realistic human anatomy principles.}
    \label{fig:skill_7}
    \vspace{-4mm}
\end{figure*}
\clearpage

\begin{figure*}[!t]
    \centering
    \includegraphics[width=\textwidth]{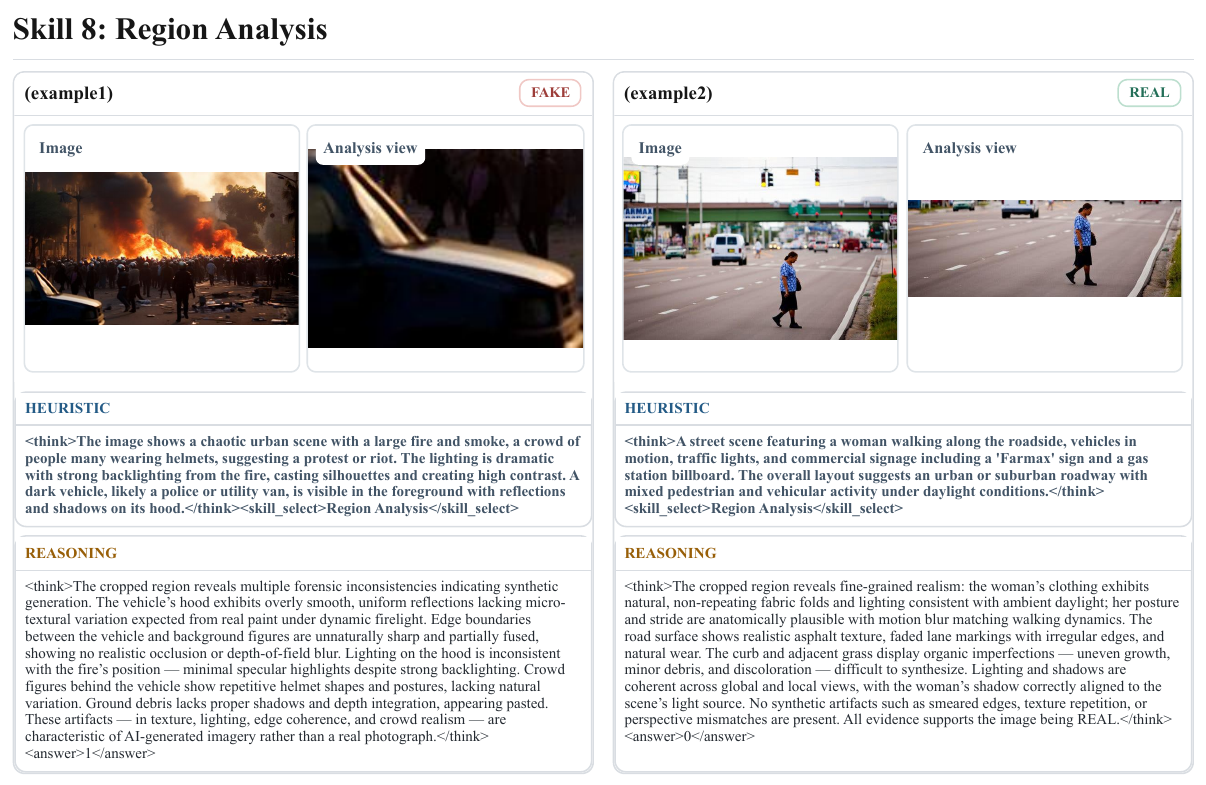}
    \caption{\textbf{Region Analysis.} Skill-based forensic analysis focusing on localized scene regions and object-level details to identify potential inconsistencies in local structures, textures, and semantic compositions.}
    \label{fig:skill_8}
\end{figure*}
\clearpage

\begin{figure*}[!t]
    \centering
    \includegraphics[width=\textwidth]{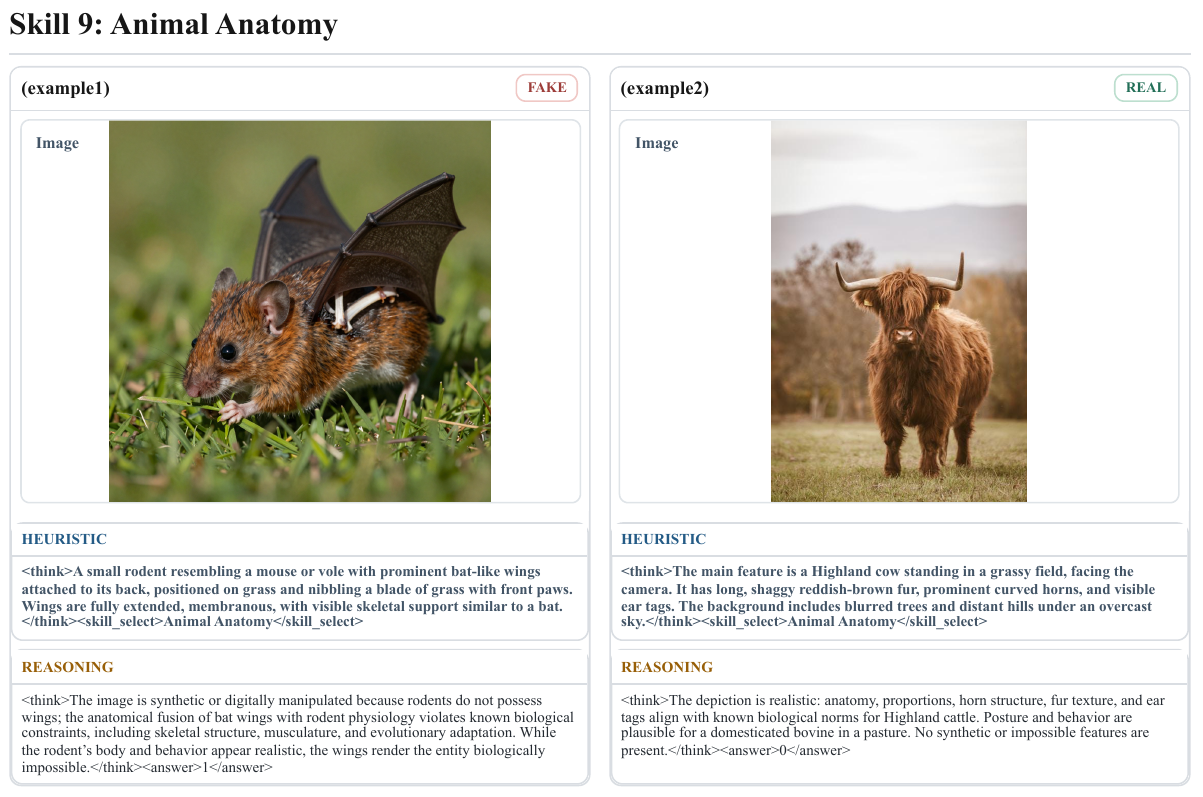}
    \caption{\textbf{Frequency Consistency.} Skill-based forensic analysis that incorporates frequency-spectrum analysis models to examine whether the image exhibits realistic frequency distributions and spectral consistency patterns.}
    \label{fig:skill_9}
    \vspace{-4mm}
\end{figure*}
\clearpage

\begin{figure*}[!t]
    \centering
    \includegraphics[width=\textwidth]{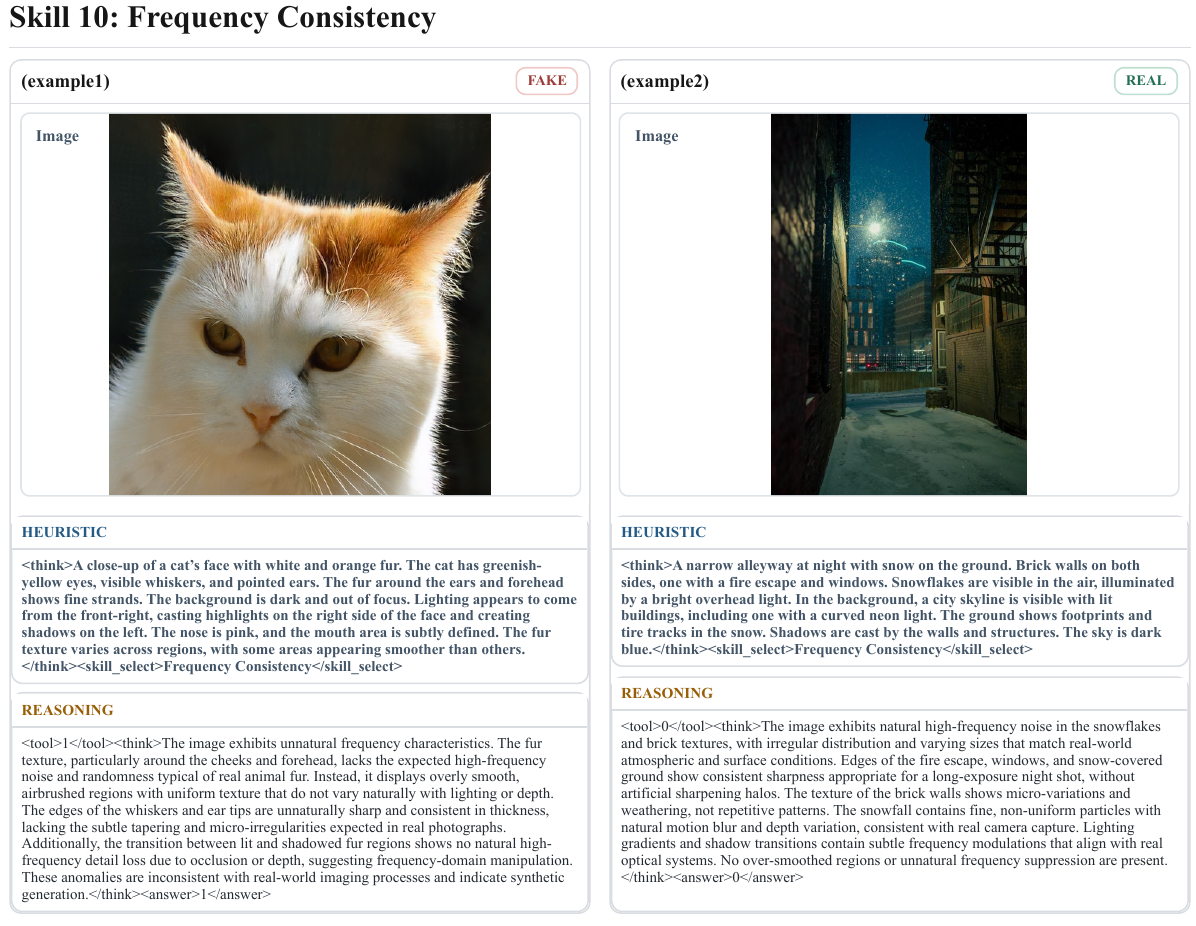}
    \caption{\textbf{Frequency Consistency.} Skill-based forensic analysis that incorporates frequency-spectrum analysis models to examine whether the image exhibits realistic frequency distributions and spectral consistency patterns.}
    \label{fig:skill_10}
    \vspace{-4mm}
\end{figure*}
\clearpage

\begin{figure*}[!t]
    \centering
    \includegraphics[width=\textwidth]{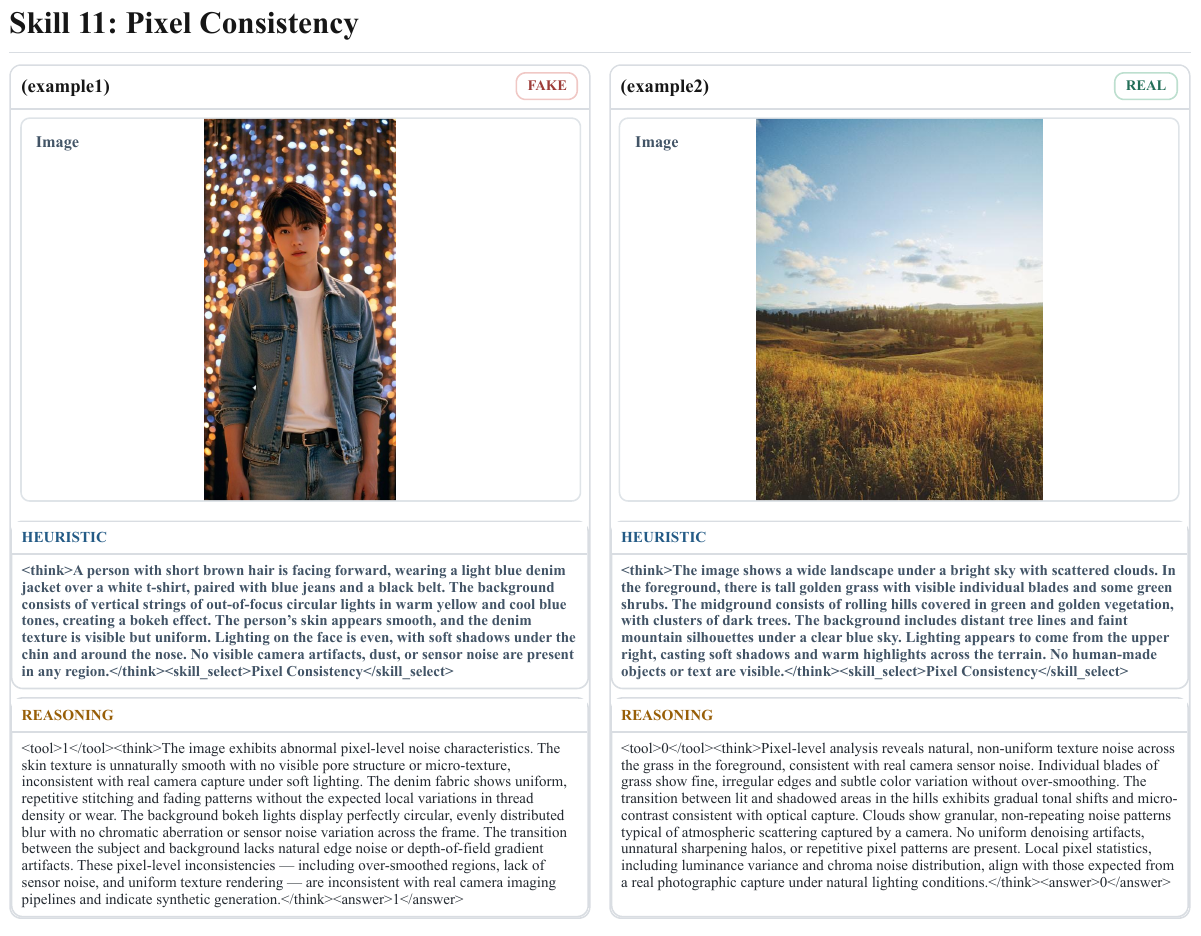}
    \caption{\textbf{Pixel Consistency.} Skill-based forensic analysis that leverages pixel-level forensic models to examine local pixel distributions, texture continuity, and low-level visual inconsistencies.}
    \label{fig:skill_11}
\end{figure*}
\clearpage

\begin{figure*}[!t]
    \centering
    \includegraphics[width=\textwidth]{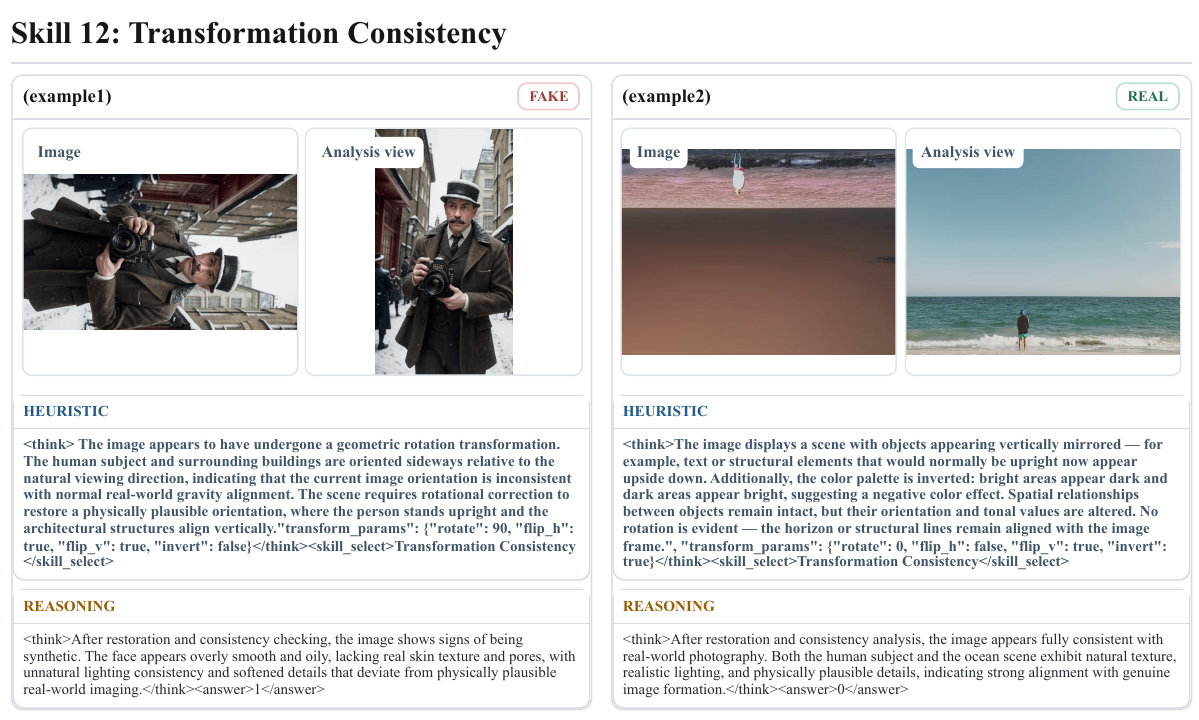}
    \caption{\textbf{Transformation Consistency.} Skill-based forensic analysis that evaluates robustness under geometric rotation and color transformations, examining whether structural layout and chromatic relationships remain consistent and physically plausible after such changes.}
    \label{fig:skill_12}
\end{figure*}

\end{document}